\documentclass[10pt]{article} 
\usepackage[accepted]{tmlr}

\usepackage{amsmath,amsfonts,bm}

\def\eqref#1{equation~\ref{#1}}

\def\1{\bm{1}}

\DeclareMathAlphabet{\mathsfit}{\encodingdefault}{\sfdefault}{m}{sl}
\SetMathAlphabet{\mathsfit}{bold}{\encodingdefault}{\sfdefault}{bx}{n}

\usepackage{hyperref}
\usepackage{url}
\usepackage{subfigure}
\usepackage{graphicx}
\usepackage{amsmath,amssymb}
\usepackage{lipsum}
\usepackage{multirow}
\usepackage{amssymb}
\usepackage{pifont}
\usepackage{booktabs}
\usepackage{algorithm}
\usepackage{etoolbox}
\usepackage{tikz}
\usepackage{mathrsfs}
\usetikzlibrary{calc,matrix}
\usepackage{enumitem}
\usepackage{xcolor}

\let\classAND\AND
\let\AND\relax
\usepackage{algorithmic}

\let\AND\classAND
\AtBeginEnvironment{algorithmic}{\let\AND\algoAND}

\floatname{algorithm}{Algorithm}

\title{TERC: A Transfer Entropy Redundancy Criterion for State Variable Selection in Reinforcement Learning}

\author{\name Charles Westphal \email charles.westphal.21@ucl.ac.uk \\
       \addr UCL Centre for Artificial Intelligence\\
       Department of Computer Science\\
       University College London\\
       Gower Street, London WC1E 6BT, United Kingdom
       \AND
       \name Stephen Hailes \email s.hailes@ucl.ac.uk \\
       \addr Department of Computer Science\\
       University College London\\
       Gower Street, London WC1E 6BT, United Kingdom
         \AND
       \name Mirco Musolesi \email m.musolesi@ucl.ac.uk \\
       \addr UCL Centre for Artificial Intelligence\\
       Department of Computer Science\\
       University College London\\
       Gower Street, London WC1E 6BT, United Kingdom \\
       Department of Computer Science and Engineering \\
       University of Bologna \\
       Viale Del Risorgimento 2, 40136 Bologna, Italy
       }

\def\openreview{\url{https://openreview.net/forum?id=J0ad21E0vX}} 

\begin{document}

\maketitle

\begin{abstract}
Identifying the most suitable variables to represent the state is a fundamental challenge in Reinforcement Learning (RL). These variables must efficiently capture the information necessary for making optimal decisions. In order to address this problem, in this paper, we introduce the Transfer Entropy Redundancy Criterion (TERC), an information-theoretic criterion, which determines if there is \textit{entropy transferred} from observable state variables to actions during training. We define an algorithm based on TERC that provably excludes variables from the observable state that do not affect the agent's policy during learning. This yields compact state representations that reduce inference time by up to $2.6\times$.
Our approach is policy-dependent, making it agnostic to the underlying learning algorithm. The efficiency gains we demonstrate arise at retraining and inference time on the reduced state.

Our method improves both retraining and inference efficiency. We demonstrate its effectiveness across three distinct algorithm classes, namely tabular Q-learning, Actor-Critic, and Proximal Policy Optimization (PPO), evaluated in a range of environments. Furthermore, to highlight the differences between the proposed methodology and the current state-of-the-art feature selection approaches, we present a series of controlled experiments on synthetic data, before generalizing to real-world decision-making tasks. We also introduce a representation of the problem that compactly captures the transfer of information from observable state variables to actions as Bayesian networks.
\end{abstract}

\section{Introduction}
\label{sec:Introduction}

The choice of an appropriate state representation remains a key design challenge of every Reinforcement Learning (RL) system. This process involves finding simplified representations of the information required to learn optimal policies. In order to achieve this, various techniques have been proposed \citep{finn2016,laskin2018,gelada2019,stooke2021}. Nevertheless, several issues can be identified with these approaches. Firstly, most of these methods fail to provide a mechanism by which the state's dimensionality can be reduced. Notable exceptions include the state-variable selection of \citet{hein2018} and the knockoff approach of \citet{ma2023sequential}.
Consequently, at the time of deployment, most RL systems process unnecessarily large amounts of data. Second, they are used in a black-box fashion, providing no insight into the final form of the simplified representation. Finally, they fail to provide a solution for determining the appropriate state history lengths when temporally extended states are required, as demonstrated in \cite{Mnih2015,Pan2017}. As a result of these shortcomings, most RL state design applications fall into the realm of intuition-guided heuristics \citep{dean1997,ortiz2018learning,reda2020learning,liu2020state}. The use of such \textit{ad hoc} methods results in state representations that incorporate non-informative variables, which might increase the training duration due to Bellman's curse of dimensionality \citep{bellman1959}. They may also lead to state representations that lack the required information, preventing the agent from learning optimal policies.

In order to provide an intuition of the problem, let us consider the classic cart pole environment\footnote{Many implementations of this environment exist, like that implemented in the now publicly maintained Gym/Gymnasium framework (\url{https://gymnasium.farama.org/environments/classic_control/cart_pole/}). In the evaluation of TERC presented in a later section, we use the original implementation by OpenAI \citep{brockman2016openai}.}.
Imagine a cart that can move left or right along a track. On top of this cart, there is a pole standing upright. The goal is to keep the pole balanced and prevent it from falling over. You can do this by moving the cart left or right to keep the pole centered. The challenge is to figure out the best way to move the cart to keep the pole balanced for as long as possible.
Typically, at time $t$, this environment is characterized by a 4-dimensional set of variables $s^t = [x^t, \theta^t_{pole}, \dot{x}^t, \dot{\theta^t_{pole}}]$, where $x$ indicates the cart’s $x$-position, $\theta^t_{pole}$ describes the angle of the pole, and $\dot{()}$ denotes the derivative with respect to time. Let us suppose that an additional random variable $v_{rand}$, which is independent of the physical system under consideration, is added to the environment. The resulting state will be $s_{extended}^t = [x^t, \theta^t_{pole}, \dot{x}^t, \dot{\theta^t_{pole}}, v_{rand}]$. $v_{rand}$ is not informative for the problem at hand and a method to define optimal representations should be able to exclude it. Failure to remove such variables will lead to unnecessary computations once deployed\footnote{Indeed, this is of critical importance given the increasing energy footprint of today's machine learning systems \citep{patterson_2022_carbon}.}. Moreover, uninformative variables can impede the learning process of iterative decision-making systems \citep{grooten2023}. 


To address these issues, we propose the Transfer Entropy Redundancy Criterion (TERC) for the selection of the minimal set of the state variables. TERC is a criterion that allows us to determine \textit{whether observable} state variables transfer entropy to actions during training \citep{schreiber2000}. 
More specifically, TERC is based on the quantification of the reduction in uncertainty of the realizations of the policy when considering the set of observable state variables with and without the variable under consideration. This makes TERC a policy-dependent method, while being agnostic to the underlying learning algorithm. If this value is bigger than zero, the actions are said to depend on this state variable, and TERC is verified. Under these circumstances, the state variable is considered informative and included in our representation. Once all the informative observable state variables have been identified, our agent can be re-trained on the lightweight state, leading to greater efficiency at the time of deployment. We will provide the reader with a formal definition of TERC in Section \ref{subsection:measure}\footnote{The code underlying the measures at the basis of TERC is available at the following URL: \url{https://github.com/c-s-westphal/terc-rl}.}. 

Before describing the method in detail, we explicitly state the twofold objective of TERC: (i) \textit{identifying a minimal sufficient state representation}, by providing a principled, information-theoretic criterion for which observable variables are necessary for the agent's learned policy; and (ii) \textit{improving retraining and inference efficiency}, by retraining or deploying the policy on the resulting lower-dimensional state, which reduces inference cost and can improve sample efficiency during retraining. 

In practical terms, we will introduce a set of methods that -in conjunction with the satisfaction of an assumption- will allow us to derive the minimal and optimal set of variables for state representation in presence of redundant and synergistic relationships. Even when this assumption does not hold, our method still identifies an optimal, though not necessarily minimal, representation. This set is constructed starting from an empty set and adding variables to it by verifying TERC.
Indeed, the core principle of our methodology is to include only the variables in the state representation that are informative. We will show that TERC can achieve this even in the presence of perfectly redundant state variables \citep{Frye2020,Kumar2020}. Consequently, we will show that it is possible to obtain an optimal subset whose realizations reduce the entropy of the actions identically to the original set of all observable state variables. Additionally, we will demonstrate that this method can serve a dual purpose: not only does it help in omitting uninformative variables from our representation, but it also facilitates investigating how variable importance varies as the agent trains, thereby enhancing the overall interpretability of the learning process.

In addition to discussing the theoretical basis of TERC, to further validate our approach from a practical point of view, we will present an extensive experimental evaluation through a selection of novel and existing environments. We will firstly apply TERC to purely synthetic data, considering different complex relationships between the observable variables. We will then introduce the `Secret Key Game', a novel environment in which an agent is tasked with learning a secret message from an observable state comprising not only `secret keys' (from which the secret message is calculated), but also superfluous `decoy keys'. The calculation of the secret message is based on a classic method for secure multiparty communication, as described in \citep{shamir1979share,blakely1979}. 
We will then consider a set of physics-based examples. We will consider
three very popular Gym environments: Cart Pole, Lunar Lander, and Pendulum. Finally, we will illustrate the versatility of TERC, applying it to strategic sequential games, in which the temporal dimension plays a key role. In particular, we will focus on the problem of learning optimal state history lengths when playing against a Tit-For-N-Tats (TFNT) opponent in the Iterated Prisoner's Dilemma.

\paragraph{Summary of contributions.} The contributions of this paper can be summarized as follows:
\begin{itemize}
\item We propose TERC, an effective criterion for the selection of state variables in reinforcement learning, discussing its theoretical foundations in detail.  
\item We analyze potential issues arising from the presence of perfect conditionally redundant variables in the observable state and we introduce algorithmic solutions to deal with them.
\item We corroborate our theoretical results experimentally by means of an extensive evaluation considering a variety of RL environments. 
\item Finally, we discuss the extent to which TERC enhances the interpretability of RL systems and its practical applications.
\end{itemize}

\section{Related Work} 
\label{sec:relatedwork}
As discussed in the introduction, the method presented in this paper is related to the field of feature selection in machine learning. The proposed solution is then applied to the problem of state representation in RL. 
Consequently, we will first review the state of the art in the area of feature selection methods before discussing the relevant work on state representation learning.

\paragraph{Feature selection.}\label{sec:featsel} Our objective is to devise a measure that assesses how informative state variables are given an agent's actions. This is conceptually similar to the idea of `feature importance' in the field of feature selection \citep{Ribeiro2016,shrikumar2017,plumb2018,Sundararajan2020,chen2020}. The exact definition of `feature importance' is an open problem \textit{per se} \citep{catav21,janssen2023}. However, it is generally agreed that its practical objective is to rank the input variables of a machine learning model based on their ability to predict the model output. For example, Shapley values, originally introduced in a game-theoretic context for allocating resources in cooperative systems \citep{Shapley1953}, have been repurposed as indicators of feature importance. One of the first examples was Multi-perturbation Shapley value Analysis (MSA) \citep{Keinan2004,cohen2007}, which entails considering features as agents which cooperate to predict model outcomes. The Shapley Additive Explanation \citep{Lunberg2017} and Shapely Additive Global Explanation \citep{Lunberg2017} models expanded on this, allowing the ranking of input features both globally (predicting all model outputs) and locally (predicting a single model output). 
Despite the success of Shapley value-based methods \citep{Apley2020, covert2020, kwon22}, both theoretical and experimental findings have shown that the presence of redundant information can invalidate these methods' results \citep{Frye2020, Kumar2020}.
To deal with these issues, in this work we draw inspiration from information-theoretic filter feature selection methods \citep{battiti94,tesmer2004, peng05,brown12,gao16,chen18}. Rather than ranking variables by feature importance, we propose the use of a `target set' that we populate with variables that lead to desirable properties. This is analogous to what is proposed by \cite{peng05,gao16,chen18}, except here we favorably derive the subset size, rather than requiring it as an arbitrary hyperparameter. \cite{brown12} proposed a method that did not suffer from such drawbacks, which has inspired subsequent works \citep{covert23,wollstadt23}. However, they fail to highlight optimal feature sets in some cases. Their methods add new features to the selected subset if they enhance the overall correlation with the target. This approach overlooks features that only become informative when considered in combination. Furthermore, such methods add features that maximize the correlation function. This function must be computed for each feature at every step, meaning the complexity of these techniques fails to scale linearly in time with respect to the number of features \citep{borboudakis2019forward,tsamardinos2019greedy,hao2021sparse}. Our method has the ability to evaluate the effects of complex relationships, such as redundancies and synergistic combinations, while remaining linear in time with respect to the number of features. In the specific RL setting, \citet{hein2018} apply AMIFS-based feature selection \citep{tesmer2004} to state variables; TERC differs by using transfer entropy to capture temporal/causal dependencies between states and actions, by handling conditional redundancy and synergy beyond pairwise interactions, and by providing theoretical guarantees of minimality. We also note the related line of work on RL-specific state-variable selection based on Sequential Knockoffs (SEEK)~\citep{ma2023sequential}, which uses knockoff features to test the relevance of individual state variables.

\paragraph{Unsupervised derivation of state representations in reinforcement learning.} \label{sec:unsup}
The Infomax principle has been used as the basis for several unsupervised representation learning techniques; it involves the maximization of the mutual information (MI) between network inputs and appropriately constrained outputs \citep{Linsker1988,bell1995}. The methods relying on the Infomax principle have been successfully used to learn representations of natural language \citep{devlin2018}, videos \citep{sun2019}, images \citep{hjelm2018,bachman2019}, and RL states \citep{finn2016,laskin2018}. For high-dimensional and continuous input/output spaces, the associated MI is not computable, and, for this reason, the application of the Infomax principle is not practically possible \citep{Song2020}. To address this issue, a variety of methods for estimating MI have been developed \citep{oord2018,belghazi2018,Poole2019}. 
The problem of defining a compact state representation has been studied primarily as an abstraction problem \citep{bean1987,lesort2018state}. The goal of state abstraction is to reduce the size of the associated state space by unifying the representation of the areas of the space that provide indistinguishable information. Examples of such techniques include bisimulation \citep{givan2003}, homomorphism \citep{Ravindran2003}, utile distinction \citep{mccallum1996}, and policy irrelevance \citep{jong2005}. A complementary line of work uses recurrent neural networks to compress histories of observables into compact quasi-Markovian state representations for partially observable MDPs~\citep{schafer2005solving, schaefer2007neural}. TERC is compatible with this perspective but operates at the level of variable selection rather than latent-state compression, and additionally derives the optimal state history length explicitly rather than learning it implicitly, as is done in \citet{duell2010}. These methods most commonly leverage MI-based abstraction principles (such as Infomax) to redefine the distributions associated with each of the original state variables, leading to new representations that preserve MI between sequential states, actions, or combinations thereof \citep{Schwarzer2020}; while concurrently disregarding irrelevant and redundant information.
In our study, we take a different approach, focusing on the conditional relationships between state variables and actions during the learning phase. This allows us to eliminate variables that are not informative with respect to agent training. Upon completion of this process, we are not required to consider the entire set of variables, in contrast to the methods previously discussed. In other words, our solution derives a compact state representation with fewer variables. Once deployed, this reduced subset requires less computation for measuring, processing, and storing. Additionally, in the case of applications that require a temporally-extended state representation (e.g., a sequence of frames for an arcade game), our method is able to derive the optimal state history length.


\section{Background and Notation}
\label{sec:back}

In the following, calligraphic symbols (e.g., ${\mathcal{X}}$) indicate sets, capital letters (e.g., $X$) represent random variables, and their realizations are denoted with lower-case letters (e.g., $x$). For a set $\mathcal{X}$ and subset $\mathcal{P} \subseteq \mathcal{X}$, we write $\mathcal{X}_{\backslash \mathcal{P}}$ to denote the set difference $\mathcal{X} \setminus \mathcal{P}$, and $\mathcal{X}_{\backslash (\mathcal{P}, \mathcal{P}')}$ as shorthand for $\mathcal{X} \setminus (\mathcal{P} \cup \mathcal{P}')$. We use Shannon's entropy, denoted $H(\cdot)$ \citep{shannon1948}, to quantify uncertainty. We write $\mathscr{P}(\mathcal{X})$ for the power set of $\mathcal{X}$. A glossary of terminology and a summary of notation are provided in Appendix~\ref{appendix:terminology}.

\paragraph{Reinforcement learning.}
A Markov Decision Process (MDP) is characterized by a tuple $\mathcal{M} = (\mathcal{S}, \mathcal{A}, R, T)$, consisting of the state space $\mathcal{S}$, action space $\mathcal{A}$, reward function $R$, and transition function $T$. An agent interacts with its environment by taking actions $a^t \in \mathcal{A}$ dependent on its current state $s^t \in \mathcal{S}$, generating trajectories $({s}^1, {a}^1, {s}^2, {a}^2, \dots, {s}^T, {a}^T)$, where each state is a vector ${s}^t = [{x}^t_{1}, {x}^t_2, \ldots, {x}^t_N]$. The agent learns a policy $\pi(a^t|s^t)$ that maximizes cumulative reward $J(\pi)=\mathbb{E}_{\pi}[\sum_{t=1}^{\infty}\gamma^t r^{t}]$, where $\gamma \in [0, 1]$ is a discount factor \citep{sutton2018}. After training, we sample state-action pairs from trajectories to derive the set of random variables $\mathcal{X} = \{X_{1}, X_{2}, \ldots, X_{N}\}$ and action variable $A$, with distributions $p(\mathcal{X}=s) = \frac{1}{T}\sum_{t=1}^T p(\mathcal{X}^t=s)$ and $p(A=a) = \frac{1}{T}\sum_{t=1}^T p(A^t=a)$.

\paragraph{Information-theoretic concepts.}
Transfer entropy (TE) quantifies how knowledge of one variable reduces uncertainty about another, analogous to a nonlinear formulation of Granger causality \citep{Granger1969,schreiber2000}. For variables $X$ and $Y$, TE is defined as $TE_{Y \rightarrow X} = H(X^t|X^{t-1}) - H(X^t|X^{t-1}, Y^{t-1})$. When state or action spaces are high-dimensional or continuous, we estimate TE via mutual information (MI) using neural estimators \citep{belghazi2018}. Two additional multivariate concepts are central to our analysis: \textit{redundancy}, which arises when multiple variables provide overlapping information about a target, and \textit{synergy}, which occurs when variables combine to provide information beyond their individual contributions \citep{williams2010}. Both phenomena require careful treatment in feature selection, as standard correlation-based methods may fail to identify redundant variables or miss synergistic relationships entirely \citep{Frye2020,Kumar2020}.

\section{Problem Statement and Training\&Data-Collection Pipeline}\label{sec:problem_statement}

\paragraph{Training and data-collection pipeline.} The TERC workflow proceeds in three stages:
\begin{enumerate}[noitemsep]
\item \textit{Initial training.} Train an agent to convergence on the full observable state $s^t=[x^t_1,\dots,x^t_N]$ (where $s^t\in\mathcal{S}$ and $x_i$ is a realization of $X_i\in\mathcal{X}$) using any standard RL algorithm, yielding a policy $\pi$.
\item \textit{Trajectory sampling.} Roll out $\pi$ to collect state--action trajectories, and from them construct the empirical distributions of $A$ and $\mathcal{X}$ (see Section~\ref{sec:back}).
\item \textit{Variable selection.} Apply TERC to identify state variables on which the agent's actions exhibit zero dependence; these are excluded to obtain $\mathcal{X}_*\subseteq\mathcal{X}$, on which a new agent can be retrained for deployment.
\end{enumerate}

\textbf{Problem statement.} We now introduce the problem statement in a formal way.
Given the set of all observable state variables, $\mathcal{X} = \{X_1, X_2\ldots X_N\}$, we aim to obtain the smallest possible subset $\mathcal{X}_* \subseteq \mathcal{X}$ from which RL agents can learn optimal policies.

Because TERC operates on distributions induced by $\pi$, the selected subset is defined \textit{with respect to $\pi$} and may therefore depend on both the training algorithm and the resulting policy.
%
%
%
Formally, this set is defined as follows\footnote{It is worth noting that this set is defined as a member of a set of subsets, due to the existence of multiple representations that satisfy the conditions introduced in the problem statement.}:
\begin{equation}
\label{eqn:problem_statement}
{\mathcal{X}}_* \in \{ \mathcal{P} \in \mathscr{P}({\mathcal{X}}) : (|\mathcal{P}| = \min\limits_{H({A}|\mathcal{P}) = H({A}|{\mathcal{X}})} |\mathcal{P}|) \quad  \& \quad (H({A}|\mathcal{P}) = H({A}|{\mathcal{X}}))\}.
\end{equation}
Since TERC only removes variables whose transfer entropy to actions is zero, the reduced state satisfies $H(A\mid\mathcal{X}_*)=H(A\mid\mathcal{X})$, so the reduced decision process retains the Markov property with respect to $\pi$ and the convergence guarantees of Theorem~4(1) in~\citet{li2006}. Additionally, diminishing the dimensionality of the state space results in a proportional reduction of the regret bounds \citep{yang20}.
Our aim is to develop an approach to derive  ${\mathcal{X}}_*$, which can then be used as the new state representation for agent training, such that $s^t_* = [x^t_1, x^t_2\ldots x^t_M]$, where $s^t_* \in \mathcal{S}_*$ and $x^t_i$ is the realization of $X_i \in \mathcal{X}_*$. 

\section{Constrained Perfect Multivariate Conditional Redundancy}
\label{sec:infomationCR}

Constrained Perfect Multivariate Conditional Redundancy (CPMCR) is a condition that, if true, signifies the existence of two disjoint subsets $\mathcal{P}, \mathcal{P}' \subseteq \mathcal{X}$ (where $\mathcal{P} \neq \mathcal{P}'$) that convey identical information about a target variable. When CPMCR holds, many existing feature importance methods fail to produce correct results \citep{breiman2001,debeer20,Frye2020,Kumar2020}. Two concrete cases illustrate this: if two observable state variables are identical (say $X_1\equiv X_2$), then $\mathcal{P}=\{X_1\}$ and $\mathcal{P}'=\{X_2\}$ form trivially redundant subsets that satisfy CPMCR. Second, if $X_1$ and $X_2$ are independent binary variables with $X_3 = X_1\oplus X_2$, then the single variable $X_3$ and the pair $\{X_1, X_2\}$ convey identical information about the XOR target. Our goal is to detect such redundant subsets and include only the smallest in our final state representation.

To formally define CPMCR, we first require that each element of a subset contributes meaningfully to its information content. For subset $\mathcal{P}'$, we define:
\begin{equation}
\label{eqn:elems_pcr_cond}
\psi_{\mathcal{P}'} = \left(\forall {P}' \in \mathcal{P}': H(Y|\mathcal{X}_{\backslash (\mathcal{P},\mathcal{P}')} \cup \mathcal{P}')  < H(Y|\mathcal{X}_{\backslash (\mathcal{P},\mathcal{P}')} \cup \mathcal{P}'_{\backslash {P}'})\right),
\end{equation}
where $\mathcal{P}'_{\backslash P'}$ denotes subset $\mathcal{P}'$ with element $P'$ removed. This condition ensures that every element of $\mathcal{P}'$ contributes to reducing uncertainty about $Y$; removing any single element increases conditional entropy. An analogous condition $\psi_{\mathcal{P}}$ applies to subset $\mathcal{P}$. 

We now define CPMCR, denoted $\Psi(Y|\mathcal{X})$, as follows:
\begin{equation}
\label{eqn:init_pcr}
\begin{split}
\Psi(Y|\mathcal{X})  =  \Big(\exists \mathcal{P}  \in \mathscr{P}(\mathcal{X}), \exists \mathcal{P}'&\in  \mathscr{P}(\mathcal{X}_{\backslash \mathcal{P}})  : \\ 
& H(Y|\mathcal{X}) = H(Y|\mathcal{X}_{\backslash  \mathcal{P}'}) = H(Y|\mathcal{X}_{\backslash \mathcal{P}})  < H(Y|\mathcal{X}_{\backslash ( \mathcal{P}, \mathcal{P}')} ) \\ 
& \land \; \psi_{\mathcal{P}'} \;\land\; \psi_\mathcal{P}\Big).
\end{split}
\end{equation}
When $\Psi(Y|\mathcal{X})$ holds, two subsets $\mathcal{P}$ and $\mathcal{P}'$ each reduce the uncertainty of $Y$ by identical amounts—removing either subset from $\mathcal{X}$ leaves the conditional entropy $H(Y|\mathcal{X})$ unchanged, yet removing both increases it. This presents a challenge for feature selection: including both subsets is redundant, but existing methods either include both \citep{catav21,jansen2022} or exclude both \citep{debeer20}.

\section{Approach}
\label{sec:approach}
In this section, we will present the design of TERC, first discussing a na\"{i}ve solution and then examining key aspects of the problem, namely perfect conditional redundancy and synergy.

\subsection{Overview}
\label{sec:overview}

To guide the reader, we briefly summarize the structure of the presentation of our approach. The main steps are outlined below:
\begin{itemize}
\item \textbf{TERC criterion (Section~\ref{subsection:measure}).} We introduce the mathematical criterion TERC uses to determine whether actions depend on particular observable state variables.
\item \textbf{Naïve solution (Section~\ref{sec:simple}).} We derive a straightforward method for identifying an appropriate representation $\mathcal{X}_*$. This method relies on the assumption that CPMCR does not hold. In practice, this assumption rarely holds, so we refer to it as the naïve solution.
\item \textbf{Practical method (Section~\ref{sec:prac_app}).} We extend the naïve approach to handle the presence of CPMCR under a weak assumption (Condition~1), yielding a practical and optimal method.
\end{itemize}
We note that the target representation $\mathcal{X}_*$ is defined as a \textit{member} of a set of subsets, since multiple representations may satisfy the conditions introduced in Section~\ref{sec:problem_statement}.

\paragraph{Practical thresholding (preview).} Throughout this section we discuss the criterion in its idealized form, namely whether the quantity $\Phi_{X_i;\mathcal{X}\to A}$ equals zero. In finite samples, exact equality to zero is not a practical test. Our implementation therefore never compares a point estimate to zero directly: instead, as detailed in Section~\ref{subsection:practical}, we compare each variable's estimated transfer-entropy contribution against a null distribution generated from a known-uninformative random variable with a $95\%$ confidence threshold.

\subsection{Design of the Transfer Entropy Redundancy Criterion (TERC)}
\label{subsection:measure}
We now introduce the mathematical details of the Transfer Entropy Redundancy Criterion (TERC).  TERC is based on the concept of transfer entropy (TE), as we are interested in how variables \textit{influence} actions using Granger's interpretation \citep{Granger1969}. Specifically, TERC quantifies the reduction in uncertainty associated with realizations of ${A}$ when considering the set ${\mathcal{X}}$ with and without a given variable. If this value is bigger than zero, the actions are said to depend on this state variable, and TERC is verified.
More formally, we define TERC as follows: 
\begin{equation}
\label{eqn:TE_measure}
    \Phi_{{X}_i;{\mathcal{X}} \rightarrow {A}}  = H({A}|{\mathcal{X}}_{\backslash {X}_i}) - H({A}|{\mathcal{X}}) > 0.
\end{equation}
$\Phi_{{X}_i;{\mathcal{X}} \rightarrow {A}}$ describes how actions depend conditionally on observable state variables; therefore, we graphically represent these directed dependencies as Bayesian networks. For a full description of how to estimate the measure defined in Equation \ref{eqn:TE_measure}, refer to Algorithm \ref{alg:te_est} in Appendix \ref{appendix:te_alg}. 

  $H({A}|{\mathcal{X}}_{\backslash {X}_i}) - H({A}|{\mathcal{X}})$ quantifies the reduction in entropy of the realizations of ${A}$, when variable ${X}_i$ is removed from the set ${\mathcal{X}}$. Positive values of this quantity indicate the actions are dependent on ${X}_i$ and, therefore, they should be maintained in the state representation. In this case, we describe $X_i$ as satisfying TERC. Conversely, if there is no reduction in the entropy of the values of ${A}$, then $\Phi_{{X}_i;{\mathcal{X}} \rightarrow {A}}= 0$\footnote{This is due to the non-negativity of the measure defined in Equation \ref{eqn:TE_measure}, a property that we prove in Appendix \ref{appendix:noneg}.} and the variable does not satisfy TERC. In this case, by definition, we can say that the information provided by variable $X_i$ is either irrelevant or redundant and removing ${X}_i$ from the state should lead to a policy that can perform inference more efficiently. It follows that we should only add variables to ${\mathcal{X}}_*$, the subset of maximal information and minimal cardinality, if they satisfy TERC ($\Phi_{{X}_i;{\mathcal{X}} \rightarrow {A}} > 0$). 
  
  

\subsection{A Na\"{i}ve Solution} \label{sec:simple}


We now describe a simple application of TERC, our previously defined criterion. To begin, we instantiate an empty set ${\mathcal{X}}_{\Phi}$. We then populate this set via the simultaneous addition of variables if their removal from set ${\mathcal{X}}$ increases the conditional entropy of ${A}$. We write this more formally as:
\begin{equation}
\label{eqn:origsubset}
 \begin{split}
{\mathcal{X}}_{\Phi} = \{{X}_i \in {\mathcal{X}} : \Phi_{{X}_i;{\mathcal{X}} \rightarrow {A}} > 0\}.
\end{split}
\end{equation}
However, in case of CPMCR, the subset ${\mathcal{X}}_{\Phi}$ does not satisfy $H({A}|{\mathcal{X}}_{\Phi}) = H({A}|{\mathcal{X}})$. We formally write this by means of the following lemma:

\noindent \textbf{Lemma 1.} \textit{If CPMCR exists in $\mathcal{X}$ (i.e., $\Psi(A|\mathcal{X})$ holds), then $H({A}|{\mathcal{X}}) < H({A}|{\mathcal{X}}_{\Phi})$.}

\noindent \textit{Proof.}  Refer to Appendix \ref{appendix:lem1pmcr}.

\noindent In Lemma 1, we show that, if $\Psi(A|\mathcal{X})$ is verified, $H({A}|{\mathcal{X}})  < H({A}|{\mathcal{X}}_{\Phi})$  and, therefore, ${\mathcal{X}}_{\Phi} \neq {\mathcal{X}}_*$. On the contrary, by assuming that there are no cases of CPMCR in ${\mathcal{X}}$, we can prove the following:

\noindent \textbf{Theorem 1.} \textit{If no CPMCR exists in $\mathcal{X}$ (i.e., $\neg\Psi(A|\mathcal{X})$), then the set ${\mathcal{X}}_{\Phi}$ defined in Equation~\ref{eqn:origsubset} satisfies ${\mathcal{X}}_* = {\mathcal{X}}_{\Phi}$.}

\noindent \textit{Proof.} See Appendix \ref{appendix:t1}.

\noindent However, $\neg \Psi(A|\mathcal{X})$ is not always satisfied.




\subsection{TERC in Practice}\label{sec:prac_app}  
In the preceding section, we presented a naive application of our criterion TERC, showing that CPMCR cases can prevent us from achieving the goal outlined in the problem statement. 
In this section, we address this issue by presenting an algorithm—valid under a weak assumption—that applies TERC in a computationally efficient manner. To begin, we motivate the need for our condition before formally introducing it and our algorithm.
 

\begin{algorithm}[!t]
    \caption{A Simple State Variable Selection Method Based on TERC}
    \label{alg:algorithm}
    \textbf{Input}: Observable state and action variables generated by sampling from learning trajectories: $\mathcal{X}$ and $A$ respectively (see Section \ref{sec:back}). \\
    \textbf{Output}: The smallest subset of ${\mathcal{X}}$ that still fully describes the agents actions: ${\mathcal{X}}_{A_1}$
    \begin{algorithmic} [1]
    \STATE Initialize ${\mathcal{X}}_{A_1} = \{\}$
    
    \FOR{$i = 1$ to $N$}
        \IF{$\Phi_{{X}_i; {\mathcal{X}} \rightarrow {A}} = 0$} 
                \STATE ${\mathcal{X}} = {\mathcal{X}} \backslash \{{X}_i\}$
        \ELSE
            \STATE ${\mathcal{X}}_{A_1} = {\mathcal{X}}_{A_1} \cup \{{X}_i\}$
        \ENDIF
    \ENDFOR
    \STATE \textbf{return} ${\mathcal{X}}_{A_1}$
    \end{algorithmic}
\end{algorithm}


Instead of adding features to our selected set simultaneously, suppose we iterate through them randomly and add them one at a time while re-verifying TERC. As a result, if two variables are perfectly redundant, the first one encountered will be removed. Upon re-verifying TERC for the second redundant variable, the removal of the first renders the second non-redundant. Consequently, we negate the issue described in the preceding section. However, if rather than consider two redundant variables we consider two redundant subsets, we have no mechanism that ensures we encounter and remove the smaller of the two subsets. Therefore, we can only be sure that we include the minimum number of variables if both redundant subsets are of equal size. This observation leads us to define the following condition:

\noindent \textbf{Condition 1.} \textit{Let $\psi_\mathcal{P}$ be as defined in Equation \ref{eqn:elems_pcr_cond} and let $\mathcal{P} \in \mathscr{P}({\mathcal{X}})$. We define the condition as follows}
\begin{equation}
\label{eqn:a1}
\begin{split}
C_1  =  (\forall \mathcal{P}\in \mathscr{P}({\mathcal{X}}), \nexists \mathcal{P}'  \in \mathscr{P}({\mathcal{X}})    : & |\mathcal{P}| \neq |\mathcal{P}'|  \quad  \& \quad  \\ & H({A}|{\mathcal{X}}) = H({A}|{\mathcal{X}}_{\backslash  \mathcal{P}'}) = H({A}|{\mathcal{X}}_{\backslash \mathcal{P}})  < H({A}|{\mathcal{X}}_{\backslash ( \mathcal{P}, \mathcal{P}')} )\quad  \& \quad \\ &\psi_\mathcal{P}\quad  \& \quad \\ &\psi_{\mathcal{P}'}) .
\end{split}
\end{equation}

In Appendix \ref{appendix:condition_1}, we show that this condition is satisfied for all of the datasets investigated in this paper. We now more formally describe the approach, which guarantees the derivation of ${\mathcal{X}}_*$, provided condition $C_1$ is true. Because the subset derived using this approach requires the use of Algorithm \ref{alg:algorithm} ($A_1$), it will be labeled ${\mathcal{X}}_{A_1}$. We write the three steps involved in this approach as follows: 
\setlist{nolistsep}
    \begin{enumerate}[noitemsep]
  \item Generate trajectories by training an agent using all observable variables as the state $s^t = [x^t_1, x^t_2\ldots x^t_N]$, where $s^t \in \mathcal{S}$  and $x^t_i$ is a realization of $X_i \in \mathcal{X}$.
  \item Iterate through variables ${X}_i \in {\mathcal{X}}$ and remove variables from ${\mathcal{X}}$ that satisfy $\Phi_{{X}_i;{\mathcal{X}} \rightarrow {A}}= 0$, otherwise add them to ${\mathcal{X}}_{A_1}$.
  \item Use this newly designed state for agent training, such that $s^t_{A_1} = [x^t_1, x^t_2\ldots x^t_N]$, where $s^t_{A_1} \in \mathcal{S}$ and $x^t_i$ is a realization of $X_i \in \mathcal{X}_{A_1}$. 
\end{enumerate} 

Notably, the complexity of Algorithm \ref{alg:algorithm} scales linearly in time with respect to the number of features. This makes it preferable to pre-existing techniques that include or exclude features that maximize some function based on the feature added at each step \citep{brown12,wookey2015regularized,gao16,borboudakis2019forward,tsamardinos2019greedy,covert23,bonetti2023causal}. We now introduce theoretical guarantees for this method. 

\noindent \textbf{Theorem 2.} \textit{Let ${\mathcal{X}}_{A_1}$ be the subset generated by Algorithm~\ref{alg:algorithm}. If Condition 1 (Equation~\ref{eqn:a1}) holds, then ${\mathcal{X}}_* = {\mathcal{X}}_{A_1}$.}

\noindent \textit{Proof.} See Appendix \ref{appendix:alg_correct}.

\noindent In practice, this condition is satisfied in many scenarios, including all of the cases studied in the upcoming experimental evaluation.  

\subsection{Methods for Approximating State Variable Inclusion Conditions}
\label{subsection:practical}

 
 Since we neurally estimate $\Phi_{{X}_i;{\mathcal{X}} \rightarrow {A}}$ instead of computing it directly, the estimates are susceptible to noise, which may lead to the inclusion of uninformative variables in our target set. Similarly to \cite{wollstadt23}, to approximate the condition $\Phi_{{X}_i;{\mathcal{X}} \rightarrow {A}}= 0$ for fluctuating estimates, we adopt a null model to which the values of $\Phi_{{X}_i;{\mathcal{X}} \rightarrow {A}}$ can be compared. This is done by introducing a random variable, $NM$, once training is complete. In theory, the actions should have no dependence on $NM$; therefore, $\Phi_{NM;\mathcal{X} \rightarrow {A}}$ should approach zero under all circumstances. Hence, variables that transfer entropy to actions are expected to deviate from the null model results. We assume normality for both the distribution of the null model and the variables in $\mathcal{X}$. We then use the methods outlined in \citep{neyman_pearson_1933} to determine which ${X}_i \in {\mathcal{X}}$ have values of $\Phi_{{X}_i;{\mathcal{X}} \rightarrow {A}}$ with a $95\%$ chance of falling outside the range described by the null model. These variables are considered to show a statistically significant deviation from the null model, and therefore satisfy $\Phi_{{X}_i;{\mathcal{X}} \rightarrow {A}} > 0$.

 In the upcoming experimental evaluation, we will present the upper bound of the null model's $95\%$ confidence interval as a red dashed line. observable state variables whose lower bound exceeds this dashed line are considered to satisfy TERC, and are included in our representation. Otherwise, they are excluded. 


\section{Experimental Evaluation}
We evaluate TERC on synthetic data and RL games of increasing complexity, concluding with the identification of optimal state-history lengths via the Iterated Prisoner's Dilemma.

\subsection{General Experimental Settings}

Unless otherwise stated, $\Phi_{{X}_i;{\mathcal{X}}\rightarrow {A}}$ is averaged over 10 runs while RL environment results are averaged over 5 runs; figures display $\pm 95\%$ confidence intervals with a red dotted line indicating the null-model threshold (Section~\ref{subsection:practical}). Hyperparameters are tuned via grid search (Appendix~\ref{appendix:impdet}). Across all environments, TERC selects consistent variables across seeds.

\subsection{Baselines}
We compare against two feature-selection baselines. The first is Ultra Marginal Feature Importance (UMFI) with optimal transport~\citep{janssen2023}, the only existing approach that resolves redundancies while scaling linearly in the number of features. The second is the widely adopted Permutation Importance (PI) algorithm~\citep{breiman2001}, which is also computationally efficient and is typically insensitive to hyperparameter tuning~\citep{probst2019}. Neither UMFI nor PI, nor the Sequential Knockoffs (SEEK) procedure~\citep{ma2023sequential}, are designed to detect synergistic interactions: their per-feature scoring contrasts cannot, even in principle, recover subsets that are informative only jointly (e.g.\ XOR-type relationships), which is why we focus on UMFI and PI as the representative practical baselines.

\subsection{Experiments on Synthetic Data} \label{appendix:synth}
\subsubsection{Motivation}
\label{sec:synthetic-experimental}

%
We first evaluate TERC against the baselines on synthetic data, to obtain controlled and easily interpretable results. Throughout this section (and in subsequent experiments), bar plots compare the per-variable feature-importance scores of TERC, UMFI, and PI. Variables are retained if their scores exceed the null-model threshold (red dashed line) with at least $95\%$ probability.

\subsubsection{Synthetic Data Generation}
We construct two synthetic datasets exhibiting different forms of CPMCR: the \textit{Four Redundant Variables dataset} (CPMCR between four individual variables) and the \textit{Two Redundant Triplets dataset} (CPMCR between two triplets of variables).

\begin{figure*}[t]
    
    \centering
    
    \subfigure{\includegraphics[width=14.5cm]{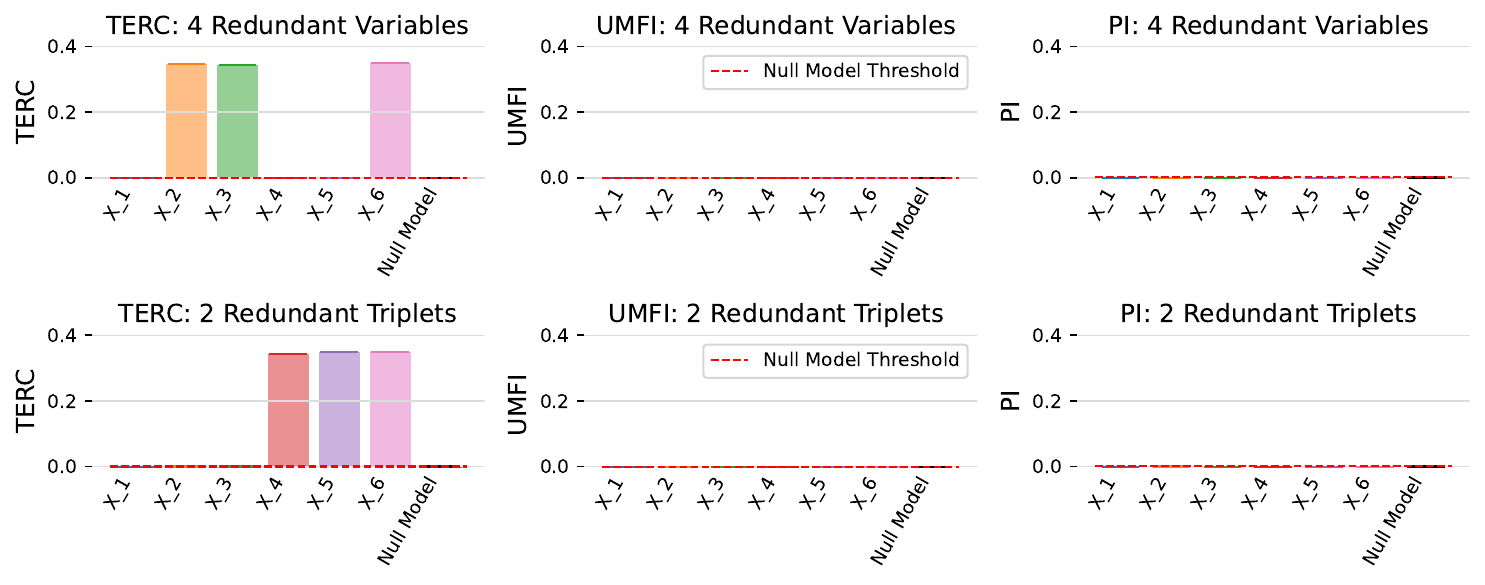} }
    \centering

    \caption{In this graph we illustrate TERC's effectiveness in dealing with complex redundancies and synergies. We plot the values for TERC, PI and UMFI for the  \textit{Four Redundant Variables} and the \textit{Two Redundant Triplets} datasets.}
    \label{fig:synth}
\end{figure*}

Both datasets use six binary variables $X_1,\dots,X_6$ of length $10{,}000$ and define the target $A$ as the three-way XOR of $X_1, X_2, X_3$. In the \textit{Four Redundant Variables dataset}, $X_4\equiv X_5\equiv X_6\equiv X_1$, so determining $A$ requires $X_2$, $X_3$, and any one of $\{X_1,X_4,X_5,X_6\}$. In the \textit{Two Redundant Triplets dataset}, $X_1\equiv X_4$, $X_2\equiv X_5$, $X_3\equiv X_6$, so CPMCR holds between any two triplets formed by selecting one member from each equivalence class; a feature-selection method should return exactly one such triplet. Both datasets combine redundancy with synergy, which pairwise methods cannot resolve (Section~\ref{sec:synthetic-experimental}).

\subsubsection{Experimental Results}
As depicted in Figure \ref{fig:synth}, for both synthetic datasets, we see that our method is able to successfully identify the minimal set needed to describe the target variable $A$. This is unlike both UMFI and PI, which fail to identify any variables, because of these methods' inability to resolve the relationships between perfectly redundant variables beyond the pairwise case.

\subsection{The Secret Key Game}
\label{sec:skg}

 \begin{figure*}[!t]%
    \subfigure(A){\includegraphics[width=9cm]{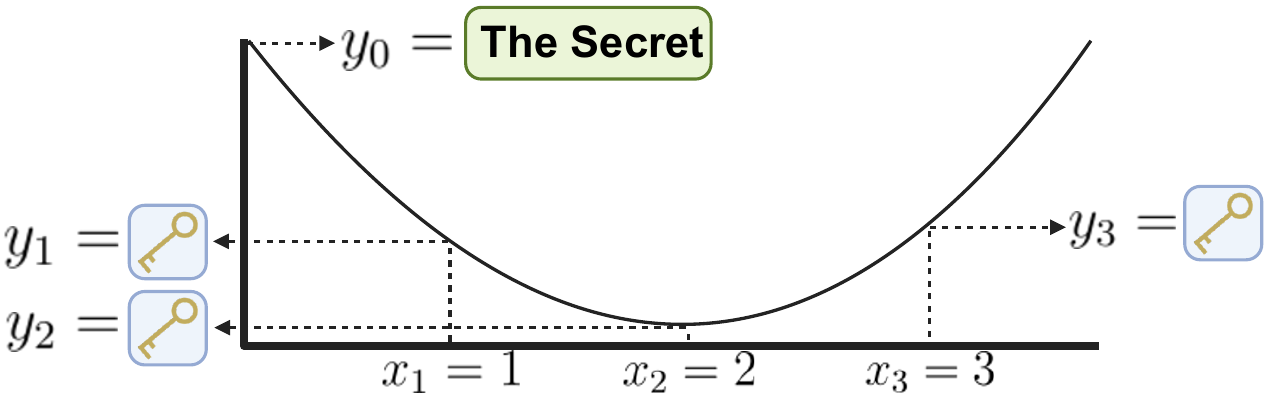} }%
    \subfigure(B){{\includegraphics[width=5.6cm]{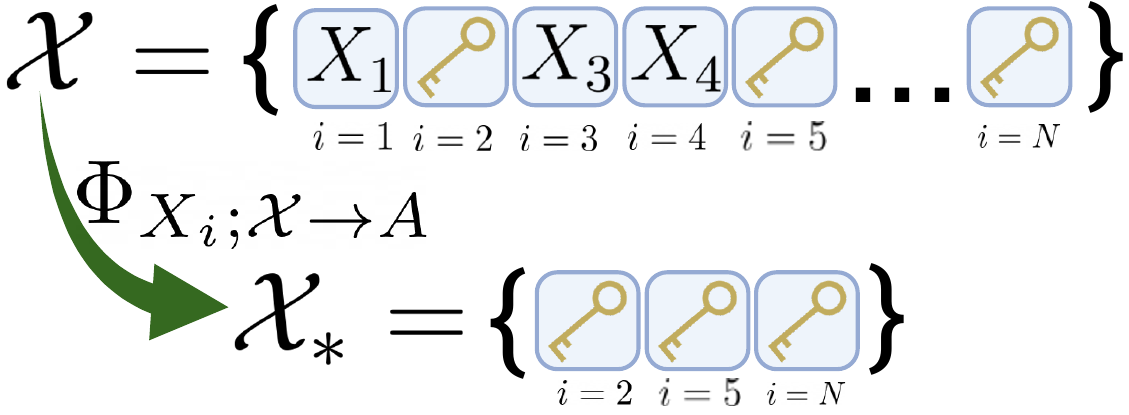} }}%
    \caption{Subfigure (A) illustrates how at least three secret keys are needed to decode a polynomial of order two in Shamir's secure multi-party communication. Subfigure (B) depicts how we use our method of state variable selection to distinguish these three secret-forming keys, from non-secret-forming keys.}
    \label{fig:skg_diag}
\end{figure*}

\subsubsection{Motivation} 

To clearly demonstrate the application and significance of the newly devised technique within the context of RL, we design a new environment, namely the Secret Key Game. The game is inspired by Shamir's secret-sharing protocol developed for secure multiparty communication. This protocol involves dividing a numerical secret into parts known as secret keys and distributing them among group members \citep{shamir1979share,blakely1979}. In this section, we introduce this protocol before adapting it to an RL game. We model the problem as follows: the agent learns to calculate the secret from a state of $N$ possible secret keys. However, only three of these keys are actually used to form the secret. The agent's ability to complete its task is consequently impeded until it learns which keys are informative.  Therefore, we can leverage our proposed method to exclude non-secret forming keys from the state, resulting in improved game performance. This improvement is attributed to the reduced dimensionality of the state space, in line with Bellman's principles \citep{bellman1959}. 

We now discuss the secret-sharing protocol that forms the basis for the game. Imagine there is a numerical secret, denoted as $y_0$, which must be kept secure until all members of a pre-determined group agree to disclose it. To accomplish this, we construct a polynomial function $f(x) = y_0 - ax + bx^2$. We then randomly generate the coefficients $a$ and $b$ in the range $[0,1]$. By doing so, we obtain a curve with a $y$-intercept equal to our secret. We then identify three points belonging to this polynomial curve, specifically $(x_1, y_1), (x_2, y_2)$, and $(x_3, y_3)$, and define the $y$-values of these points ($y_1, y_2$, and $y_3$) as our secret keys to be shared among the group members. Figure \ref{fig:skg_diag}.A visualizes this process, showing a second-order polynomial curve with three secret keys defined by the $y$-values of three distinct points, while the secret is represented by the $y$-intercept. The original secret can be derived using polynomial interpolation, but if, and only if, all three secret keys are made available.
We adapt this method of multi-party communication into a simple RL game by considering the following question: given you had $N$ keys, of which only three were being used to form $y_0$, how many incorrect guesses would it take for an RL agent to learn how to correctly calculate the secret?

The Secret Key Game is played as follows: for each iteration of the game, there is a unique new secret, which is divided into three secret keys. These keys are then hidden among $N-K$ decoy keys to form set $\mathcal{X}$, as illustrated in Figure \ref{fig:skg_diag}.B. In our experiments, $K=3$. The set $\mathcal{X}$ then serves as the state of the RL agent, considering the following reward function: $r = -|a^t-y_0|$, i.e., the negative absolute difference between the agent's action at time $t$ and the secret. 
We design the game as described to showcase the effectiveness of the state variable selection method in detecting non-secret forming keys (i.e., keys that are not inputs of the functions used to generate the secret) in $\mathcal{X}$, resulting in their removal to form ${\mathcal{X}}_*$ as illustrated in Figure \ref{fig:skg_diag}.B. This enhances the accuracy of the RL agent in approximating the $y$-intercept of the polynomial function, resulting in improved game performance.

Concretely, with observable state length equal to $25$ (three secret keys at randomly chosen indices, e.g., 2, 6, 25, and 22 decoy keys), each iteration draws a fresh observable state $s^t$ consisting of $25$ integers in $[0,10]$. The three secret-key values $y_1, y_2, y_3$ together with $x \in \{1, 2, 3\}$ define the interpolating polynomial whose $y$-intercept is the secret $y_0$. The agent's reward is $r^t = -|a^t - y_0|$ with action space $\mathcal{A}=[-40,40]$, and the transition function assigns equal probability across states ($T:\mathcal{S}\mapsto\Delta(\mathcal{S})$). The secret-key indices remain fixed across iterations of a given game.

\subsubsection{Experimental Settings} 

%
In our experiments, we consider states formed of $25$ or $50$ keys when generating trajectories. By assuming that the $y$-values of the secret keys are integers in the range $[0..10]$; the resulting secret is also an integer within the range of $y_0 \in [-40..40]$. Therefore, to ensure that the agent action space is the same as the range of the secret, we employ an RL algorithm with an 80-dimensional discrete action space $\mathcal{A} = [-40..40]$. Due to the large observable state spaces, we use a one-step temporal difference learning Actor-Critic algorithm \citep{sutton2018}, as described in \ref{sec:prac_app}. Finally, as far as the selection of the hyperparameters for the RL algorithm is concerned, we refer to Appendix \ref{appendix:impdet}.
\begin{figure*}[!t]%
    \centering
    \subfigure(A){\includegraphics[width=14.5cm]{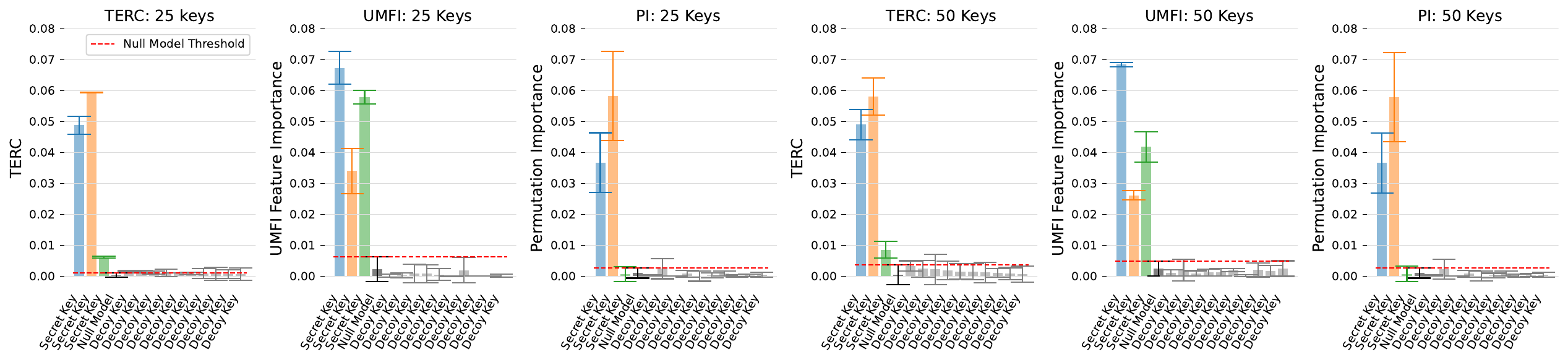} }%
    \centering
    \subfigure(B){\includegraphics[width=4cm]{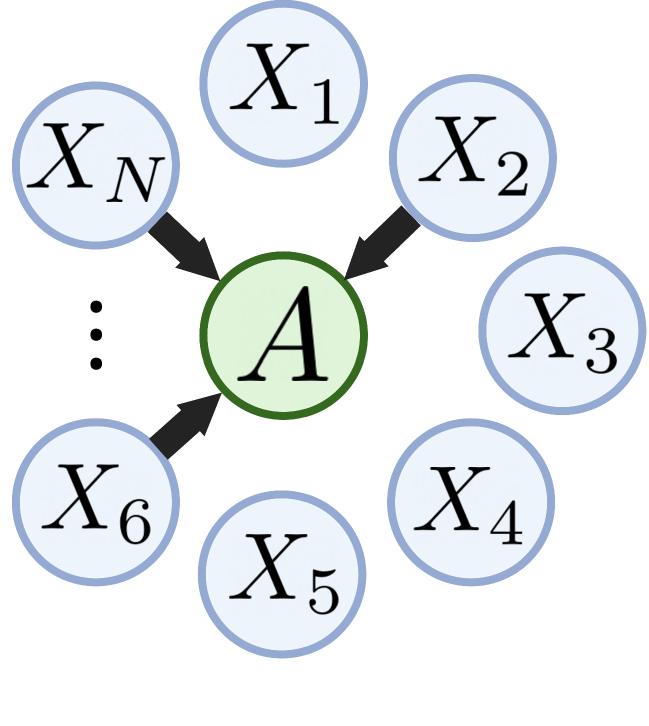} }%
    \subfigure(C){{\includegraphics[width=6cm, height=4.5cm]{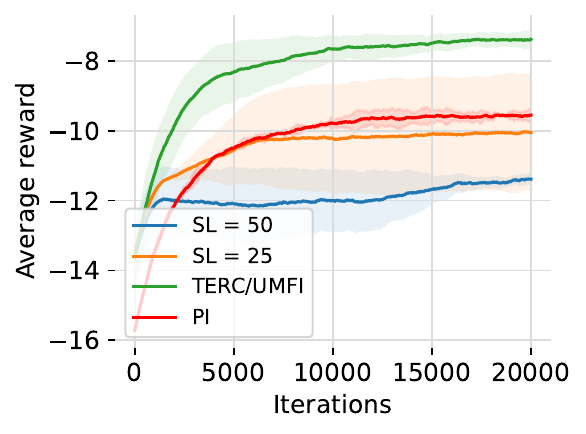} }}%
    \centering
    \caption{Subfigure (A) depicts the final feature importance values for each key in the secret game when using TERC, UMFI or PI. Subfigure (B) depicts the Bayesian network representation of TERC in the Secret Key Game if the secret keys were at index two, six and $N$. Finally, graph (C) represents how the agent training efficiency varied as a function of state length.}
    \label{fig:skg}
    
\end{figure*}

\begin{figure*}[!t]%
    \centering
    \subfigure(A){{\includegraphics[width=14.5cm]{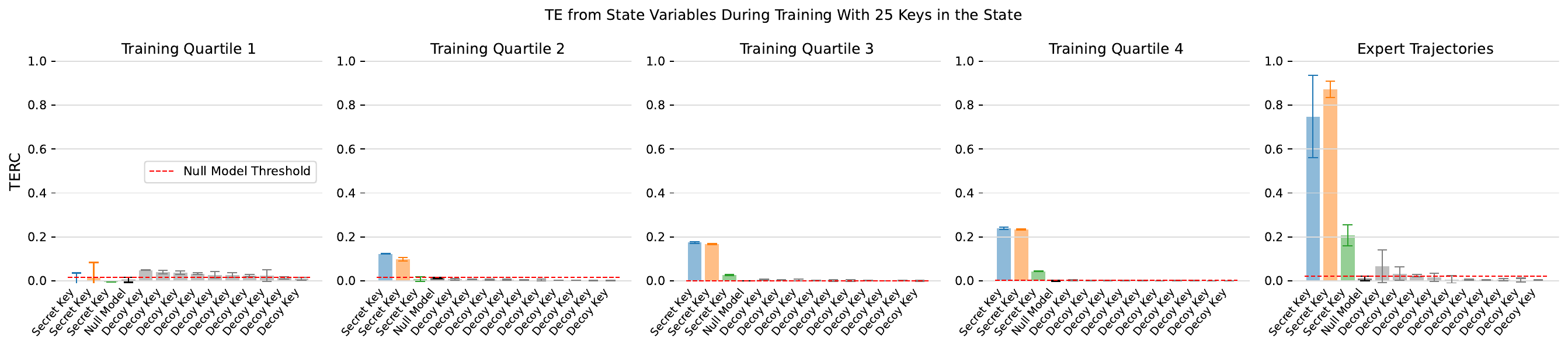} }}%
    \centering
    \subfigure(B){{\includegraphics[width=14.5cm]{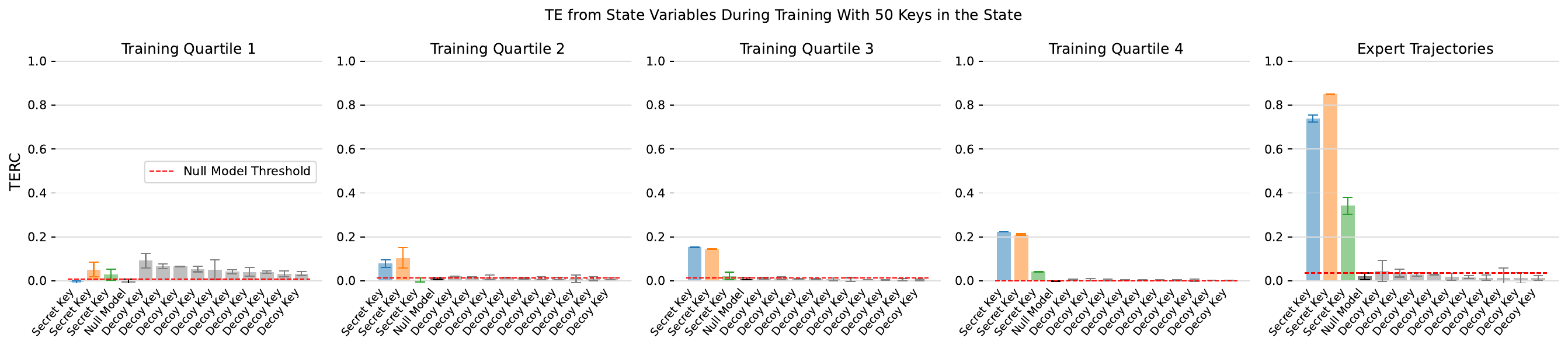} }}%
    
    \centering
    \caption{Graphs (A) and (B) show the final values computed when verifying TERC as evaluated during different training quartiles, for 25 and 50 potential secret keys. Similarly to Figure \ref{fig:skg}, we have included only TERC values for the 10 decoy keys that transferred the most entropy to the actions. }
    \label{fig:skgquart}
    \newpage
\end{figure*}

\subsubsection{Experimental Results} 
Figure \ref{fig:skg}.A plots the values of $\Phi_{{X}_i;{\mathcal{X}} \rightarrow {A}}$ that are calculated when aiming to verify TERC. Furthermore, in this plot, we also display values of UMFI and PI for each key. In Figure \ref{fig:skg}.A, there are 25 potential keys in the state, of which only three form the secret, which we label as secret keys, while the remainder have been labeled as decoy keys. According to the results of TERC in Figure \ref{fig:skg}.A, the actions were only conditionally dependent on the three secret forming keys, which were randomly selected at the start of the game as keys $2, 6 $ and $25$.  
We represent the results using a Bayesian network as illustrated in Figure \ref{fig:skg}.B. UMFI similarly leads to a set consisting of all three secret keys when applied to this simple game. However, PI can only identify two out of three secret keys. We observe similar results in \ref{fig:skg}.A when the observable state is composed of 50 keys. Figure \ref{fig:skg}.C shows how failing to implement state variable selection can deteriorate the agent performance while playing the Secret Key Game. We observe that the state designed using TERC/UMFI achieves the greatest cumulative reward. Beyond training performance, the reduced state representation also improves deployment efficiency. Table~\ref{tab:skg_inference} reports inference times for policies trained on full versus TERC-selected states. The substantial dimensionality reduction (88--94\%) yields speedups of up to $2.6\times$ on CPU for SKG-50.

\begin{table}[ht]
\centering
\caption{Inference speedup for the Secret Key Game on CPU caused by using the state selected by TERC, averaged over 5 seeds (10,000 inference calls each). $B$ denotes batch size.}
\label{tab:skg_inference}

\begin{tabular}{lcc}
\toprule
\textbf{Environment} & $B{=}1$ & $B{=}64$ \\
\midrule
SKG-25 ($25 \rightarrow 3$) & $1.43\times$ & $1.47\times$ \\
SKG-50 ($50 \rightarrow 3$) & $2.60\times$ & $2.20\times$ \\
\bottomrule
\end{tabular}

\end{table}

Figure \ref{fig:skgquart}.A and \ref{fig:skgquart}.B depict how $\Phi_{X_i;{\mathcal{X}} \rightarrow {A}}$ evolves during different stages of training when the state comprises either 25 or 50 keys. As the agent learns, it receives higher rewards by enhancing its ability to decipher the secret. This leads to the agent's actions increasingly depending on the secret keys, as shown in Figures \ref{fig:skgquart}.A and \ref{fig:skgquart}.B.

\subsection{Gym Physics Environments}
\label{sec:Gym}

\subsubsection{Motivation}
In order to show the potential of the proposed method for physics-based RL problems, we use three environments provided by OpenAI Gym \citep{brockman2016openai}, namely Cart Pole, Lunar Lander, and Pendulum.

Following \citep{grooten2023}, we `dope' the environment-provided state with a set of random variables $V_{rand_i}$. For example, in Cart Pole\footnote{\url{https://gymnasium.farama.org/environments/classic_control/cart_pole/}.} the observable state $s^t = [x^t, \dot{x}^t, \theta_{pole}^t, \dot{\theta}_{pole}^t]$ is augmented with three noise variables to give $\mathcal{X} = \{X, \dot{X}, \Theta_{pole}, \dot{\Theta}_{pole}, V_{rand_1}, V_{rand_2}, V_{rand_3}\}$. We train RL agents on the doped observable state until convergence, then verify that TERC removes $V_{rand_i}$ to recover ${\mathcal{X}}_*$, and that retraining on ${\mathcal{X}}_*$ improves learning efficiency. Examining $\Phi_{X_i;{\mathcal{X}} \rightarrow {A}}$ across training stages additionally reveals distinct phases in the behavioral dynamics of the agent, highlighting TERC’s potential for interpretability.

\subsubsection{Experimental Settings} 
For the environments with discrete action and state spaces, namely, Cart Pole and Lunar Lander, we train the agents using a one-step temporal difference Actor-Critic architecture \citep{sutton2018}. Instead, for the continuous action space environment we take into consideration, Pendulum, we use PPO \citep{schulman2017}. For more details on these methods, including hyperparameters, please refer to  Appendix \ref{appendix:impdet}.

\begin{figure*}[t]
    \centering
    \subfigure{\includegraphics[width=14.5cm]{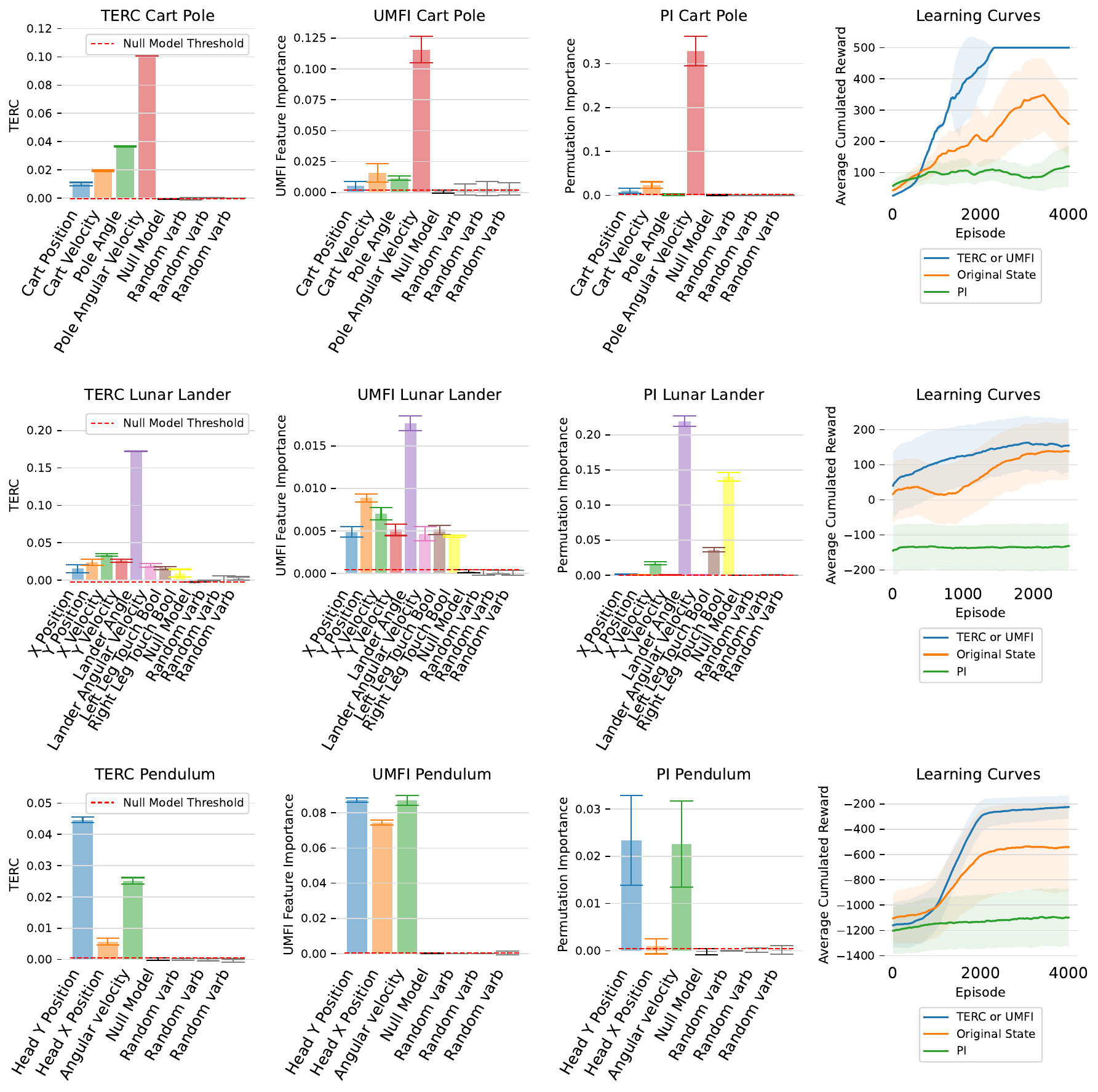} }
    \centering
    \caption{This subfigure depicts the final values obtained for $\Phi_{{X}_i;{\mathcal{X}} \rightarrow {A}}$, UMFI, and PI for Cart Pole, Lunar Lander, and Pendulum. On the right-hand side, we illustrate how failing to remove these random variables from the state of the Cart Pole playing agent degrades the performance of the game. }
    \label{fig:CartPole}
    
    \label{fig:lunar_lander}
    
    \label{fig:pend}
\end{figure*}

\begin{figure*}[t!]
    \centering
    \subfigure{\includegraphics[width=14.5cm]{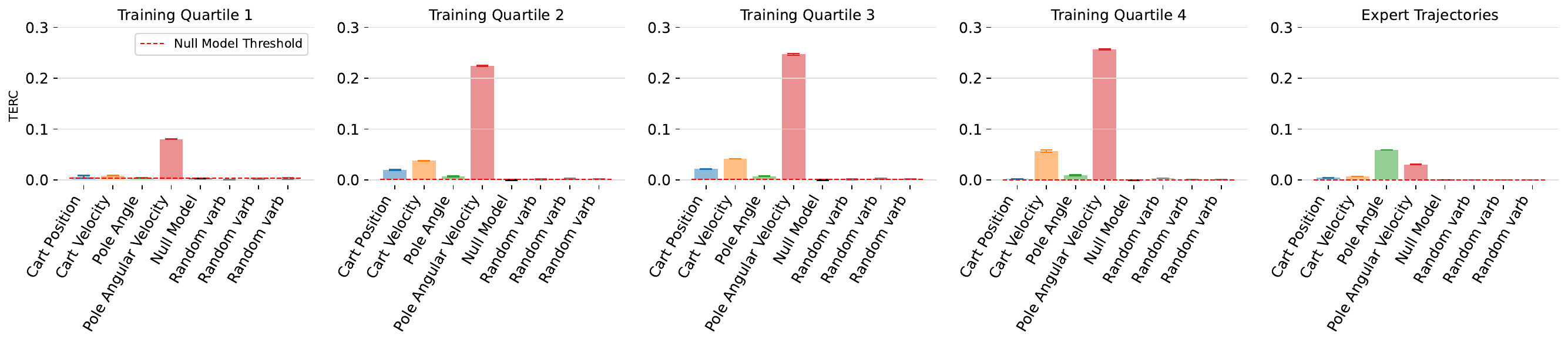} }
    \centering
    \caption{Entropy transferred from the observable state variables to the actions during different phases of training in the Cart Pole environment.}
    \label{fig:CartPole2}
    \centering
    \subfigure{\includegraphics[width=14.5cm]{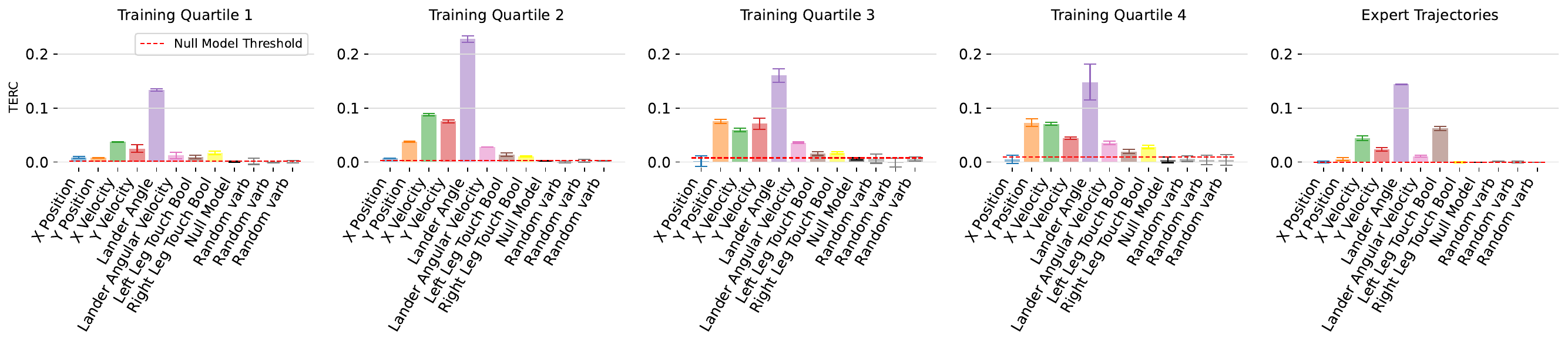} }
    \caption{Entropy transferred from the observable state variables to the actions during different phases of training in the Lunar Lander environment.}
    \label{fig:lunar_lander2}
    \centering
    \subfigure{\includegraphics[width=14.5cm]{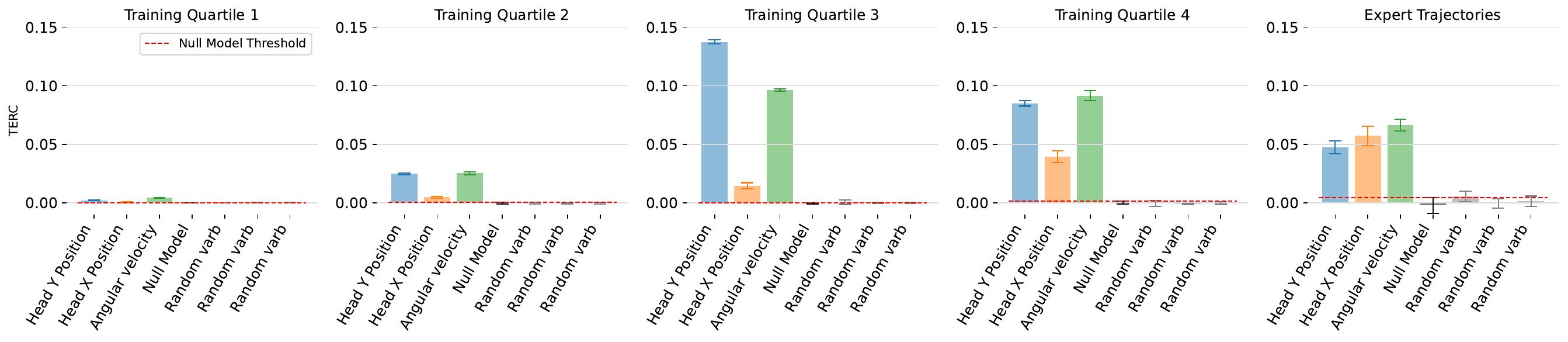} }
    \caption{Entropy transferred from the observable state variables to the actions during different phases of training in the Pendulum environment.}
    \label{fig:pend2}
\end{figure*}

\subsubsection{Experimental results} The $\Phi_{{X}_i;{\mathcal{X}} \rightarrow {A}}$ and UMFI values of the completely random variables ($V_{rand_i}$), as depicted in Figure \ref{fig:CartPole}, fall within the range of the null model. As a result, these random variables are excluded from the set $\mathcal{X}$, to form $\mathcal{X}_*$. Consequently, this updated set contains only the original variables.
PI instead fails to detect all the informative variables. Training the agent using the sets identified using TERC, UMFI and PI leads to the learning curves seen on the right-hand side of Figure \ref{fig:pend}. These graphs show how training can be sped up when using the optimal set of variables as identified using TERC and UMFI. Table~\ref{tab:gym_inference} reports that these reduced representations also yield modest inference speedups ($1.05$--$1.33\times$ on CPU), with gains scaling proportionally to the degree of dimensionality reduction.

\begin{table}[ht]
\centering
\caption{Inference speedup for Gym environments on CPU, averaged over 5 seeds (10,000 inference calls each). $B$ denotes batch size.}
\label{tab:gym_inference}

\begin{tabular}{lcc}
\toprule
\textbf{Environment} & $B{=}1$ & $B{=}64$ \\
\midrule
CartPole ($7 \rightarrow 4$) & $1.16\times$ & $1.33\times$ \\
LunarLander ($11 \rightarrow 8$) & $1.10\times$ & $1.25\times$ \\
Pendulum ($6 \rightarrow 3$) & $1.05\times$ & $1.03\times$ \\
\bottomrule
\end{tabular}

\end{table}

Now, we examine how $\Phi_{X_i;{\mathcal{X}} \rightarrow {A}}$ changes as the agent trains, demonstrating TERC's potential use for model interpretability. Figure \ref{fig:CartPole2} illustrates these changes as the agent learns the cart-pole game. Initially, the agent learns to maintain the pole's balance by moving in one direction according to the sign of the angular velocity until it reaches the boundary. As a result, during the first quarter of the training, we observe that the agent's actions depend only on the angular velocity. It is possible to observe that the agent tries to avoid the environment's edge during the second quartile of training, which leads to an increase of the value of $\Phi_{X_i;{\mathcal{X}} \rightarrow {A}}$ for the variables associated with cart velocity and $x$-position. As the agent moves into the fourth quartile, it is increasingly able to remain within the central area of the environment, reducing the need for correcting the cart position; as a result, its actions exhibit less dependence on the cart position variable. In the final graph of Figure \ref{fig:CartPole2} (right), the agent has perfected its ability to stay in the center of the environment, making only minor movements to re-adjust the angle of the pole if it deviates from the upright position. This shift in strategy leads to actions that show less dependence on the angular velocity, velocity, and position while exhibiting a stronger dependence on the angle. 

Figure \ref{fig:lunar_lander2} instead shows how an RL agent's strategy evolves while learning to play Lunar Lander. Initially, during the first quarter of training, the agent discovers the necessity of taking actions that are dependent on the lander's angle to maintain its upright position. After achieving this, the agent begins to learn to fly around the environment, leading to its actions becoming more dependent upon the speed and position variables throughout the second and third quarters of training. Finally, throughout the trajectories collected from the last quarter of training and the expert trajectories, the agent masters landing within the target area, which results in the actions depending more on the Boolean variable that describes whether the lander's legs have touched down successfully.

Figure \ref{fig:pend2} demonstrates the changes in $\Phi_{X_i;{\mathcal{X}} \rightarrow {A}}$ estimates as the agent learns how to play the Pendulum game. During the first quartile, the agent's actions are primarily stochastic due to the lack of a learned strategy, resulting in negligible $\Phi_{X_i;{\mathcal{X}} \rightarrow {A}}$ values across all variables. In the second and third quartiles, the agent learns to swing the pendulum to the upright position, making its actions increasingly dependent on the $y$-position of the pendulum head and the angular velocity, as depicted in the second and third panels of Figure \ref{fig:pend2}. Once the agent has learned to swing the pendulum upright, it then learns to keep it there by making minor side-to-side adjustments. This is reflected in the values of $\Phi_{X_i;{\mathcal{X}} \rightarrow {A}}$ reported in the third and fourth quartiles, where we observe a rising dependency on the $x$ position value of the pendulum head.

\subsection{Tit-For-N-Tats Strategy in the Iterated Prisoner's Dilemma}
\label{sec:ipd}
\subsubsection{Motivation} 
\begin{figure*}[!t]
    \centering
    \subfigure(A){\includegraphics[width=8.2cm]{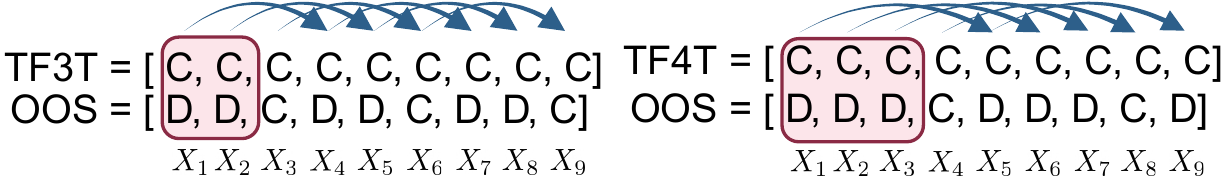}}
    \subfigure(B){\includegraphics[width=4.9cm, height = 1.4cm]{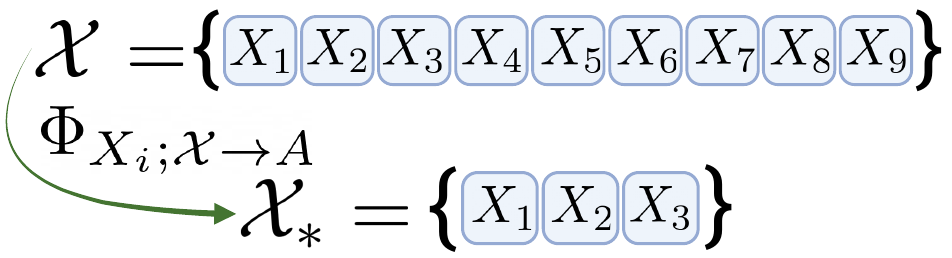}}
    \caption{In Subfigure (A) we depict the moves made by both players when playing optimally against TF3T and TF4T strategies in the Iterated Prisoner's Dilemma. The colored boxes indicate the minimum history needed to learn this optimal strategy, while the colored arrows indicate one period of a repeating action sequence, where $C$ stands for cooperative moves, while $D$ represents defective ones. In Subfigure (B), we illustrate schematically how our measure is used to determine the optimal state history length of three, out of a maximum of nine, when facing a TF4T opponent.}
    \label{fig:pd_g}
\end{figure*}

Our final environment is the Iterated Prisoner's Dilemma \citep{axelrod1981}, which we use to demonstrate that TERC can discover optimal history lengths when temporally extended states are required. The state $s^t \in \mathcal{S}$ consists of the past $l$ actions of both the player and their opponent, with each action $a^t_i \in \{C,D\}$ denoting cooperation or defection \citep{Anastassacos2020}. The reward matrix is given in Appendix \ref{appendix:pdmat}. When both contestants are maximizing reward under this matrix, $l=1$ suffices for optimality \citep{press2012iterated}; however, against other opponent policies a longer history may be required. A canonical case is Tit-For-N-Tats (TFNT): the opponent defects only after $N$ consecutive defections and otherwise cooperates. The optimal counter-strategy takes $N-1$ defective actions before cooperating, which requires a minimum history length of $N-1$ (red boxes in Figure \ref{fig:pd_g}.A, showing the optimal opponent strategies against TF3T and TF4T). We demonstrate below that TERC reliably identifies this optimal history length.

\subsubsection{Experimental Settings} 

Let $l$ denote the maximum history length of interest. We train a tabular Q-learning agent (hyperparameters in Appendix \ref{appendix:impdet}) on a state of history length $l$ against a TFNT opponent, and form ${\mathcal{X}} = \{{X}_1 \dots {X}_l\}$ from the resulting action-pair variables (Figure \ref{fig:pd_g}.A). We set $L=9$ to cover strategies up to Tit-For-10-Tats. Because the optimal counter-strategy is cyclic with period $N$, the realizations of $X_i$ and $X_{i+N}$ coincide, yielding CPMCR both between the pair $\{X_i, X_{i+N}\}$ ($\Psi_{{X}_i, X_{i+N}}(A|\mathcal{X})$) and between the full-period subsets $\mathcal{P} = \{X_i, X_{i+1}, \dots, X_{i+N-1}\}$ and $\mathcal{P}' = \{X_{i+N}, X_{i+N+1}, \dots, X_{i+2N-1}\}$ (colored arrows in Figure \ref{fig:pd_g}.A). This is the only non-synthetic experiment in which CPMCR arises in our empirical evaluation. Network hyperparameters for Algorithm \ref{alg:algorithm} are given in Appendix \ref{appendix:impdet}.

\begin{figure*}[!t]
    \centering
    \subfigure(A){\includegraphics[width=14.5cm]{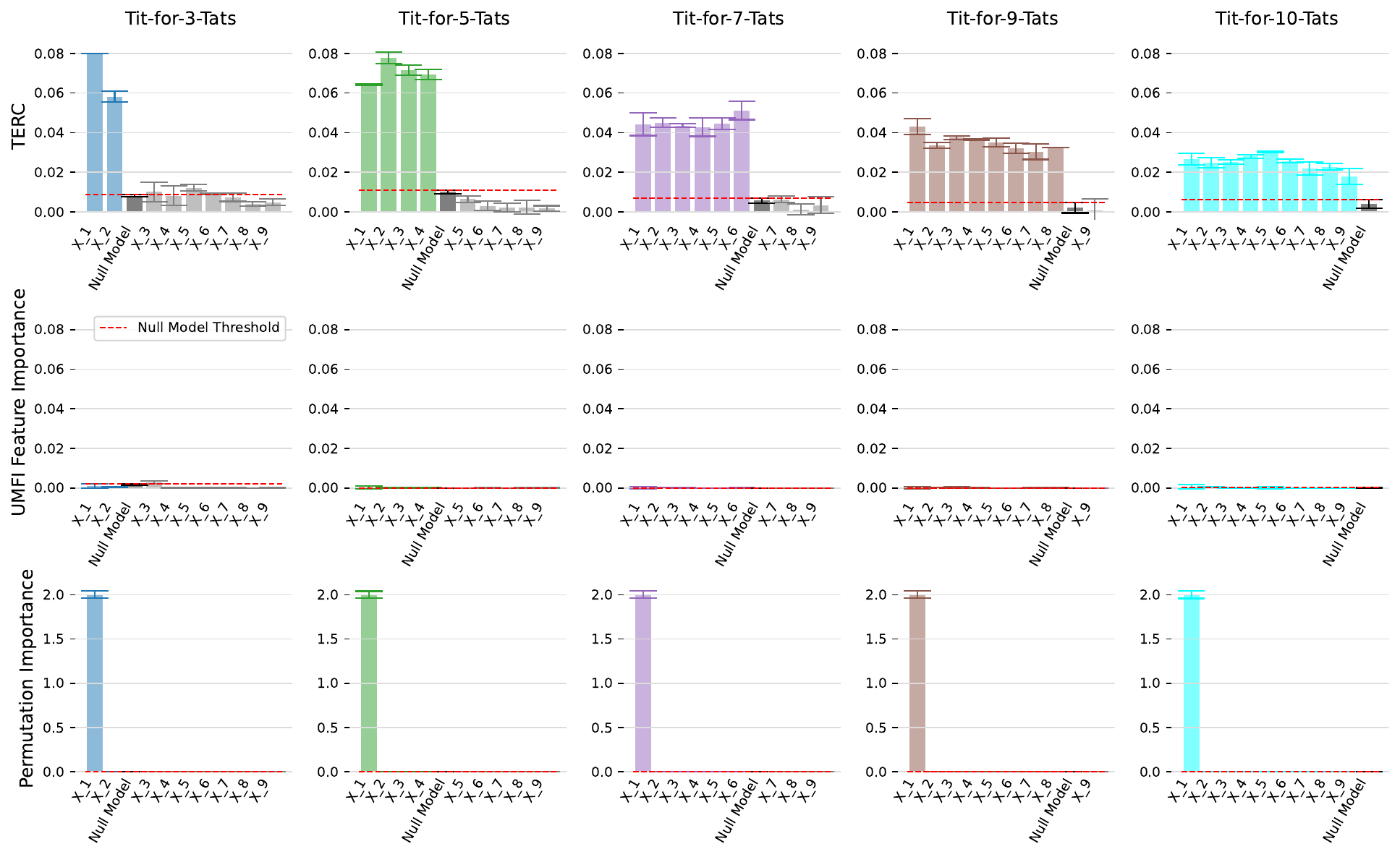} }
    \centering
    
    \subfigure(B){\includegraphics[width=3.8cm]{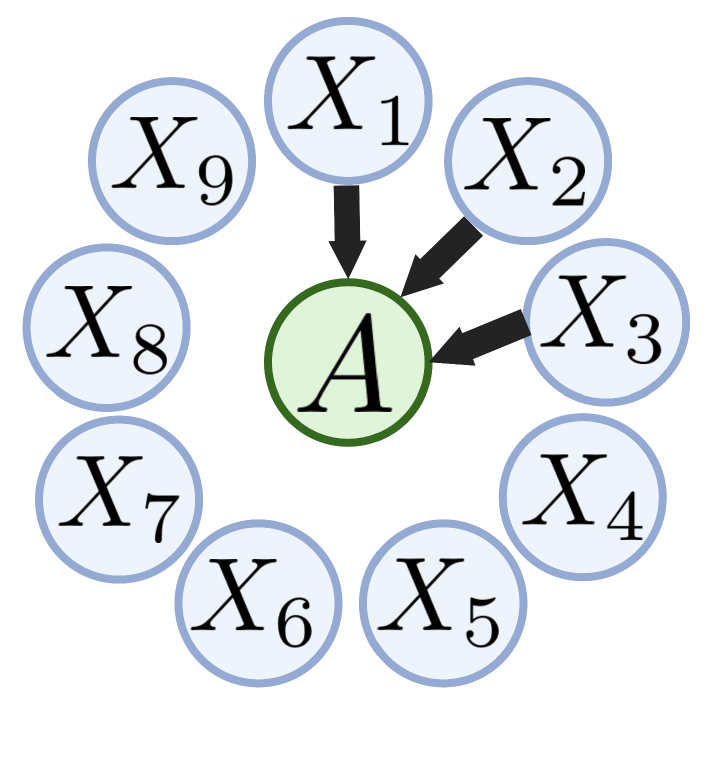} }
    \centering
    \subfigure(C){\includegraphics[width=5.5cm]{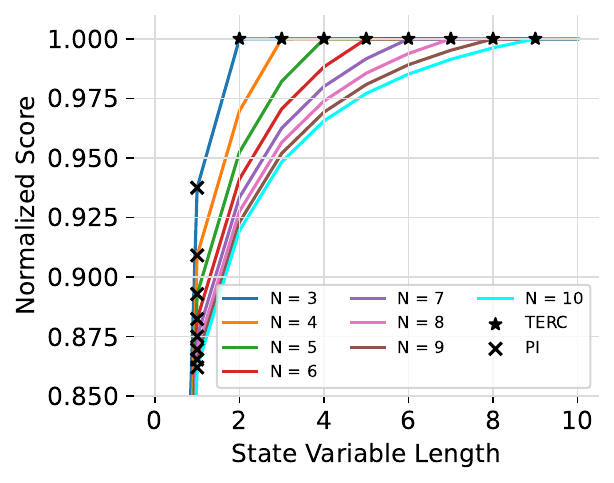} }%
    \centering
   
    \caption{In Subfigure (A) we compare $\Phi_{{X}_i;{\mathcal{X}} \rightarrow {A}}$, UMFI and PI, values for all state variables from $l=1$ to $l=9$, when playing optimally against a TFNT opponent. For clarity of presentation, we only present the results with the values of $N$ displayed above.
    These results were representative of how our algorithm performed for all TFNT opponents. Subfigure (B) shows the resulting Bayesian network representation of how the actions depend on observable state variables against a TF4T opponent.  Subfigure (C) shows how the change of the state size affects an agent's performance against a TFNT strategy. We use stars  to denote the optimal state length highlighted by our method, whereas we use crosses to represent the state length generated using PI. Finally, we omit the results for UMFI in Subfigure (C) for clarity.  }
    \label{fig:pd}
\end{figure*}

\subsubsection{Experimental Results} 
In Figure \ref{fig:pd}.A, we show the values of $\Phi_{{X}_i;{\mathcal{X}} \rightarrow {A}}$, UMFI and PI when applied to the trajectories generated in the case of a Q-learner play against a TFNT opponent. In these datasets we do not have simple pairwise redundancy and, for this reason, the assumptions in \citep{janssen2023} are not verified. Therefore, this method is not able to identify the correct set of variables; PI is similarly ineffective.  Only TERC has the ability to identify the correct set of variables. We represent the values in the TF4T case pictorially as the Bayesian network shown in Figure \ref{fig:pd}.B. This is further reflected in the agent performance results reported in Figure \ref{fig:pd}.C, where we plot the normalized cumulative reward achieved over the last $1000$ iterations of the game, as a function of the length of the state history. Since optimal play yields the maximum attainable reward, this normalized score is equivalent to the fraction of times the agent plays optimally during the last 1000 iterations of training: a value of 1 indicates that the agent is playing optimally throughout, a value of 0.9 implies it plays optimally $90\%$ of the time, and so on.
Agents trained with a history length smaller than $N-1$ converge to suboptimal strategies. Consequently, we observe that only state lengths equal to 9 achieve optimal cumulative rewards against the TF10T opponent, whereas state lengths of $2$ play optimally against the TF3T opponent. The variables that satisfy TERC in Figure \ref{fig:pd}.A can therefore be interpreted as indicating the minimum subset of ${\mathcal{X}}$ required to play optimally, corroborating the theoretical observations reported earlier. The detailed learning dynamics are reported in the Appendix \ref{appendix:tft_graphs}. TERC selects consistent state variables across seeds. Inference improvements are not reported because we employ tabular Q-learning.

\section{Conclusions}
In this article, we have introduced a novel information-theoretic methodology for state variable selection in RL, based on TERC, a criterion that can be used to verify the existence of dependencies between observable state variables and the agent's actions. The objective is the definition of the minimal subset of state variables from which an agent could learn optimal policies. We have defined this optimal subset as one whose realizations reduce the entropy of the actions identically to those of the set of all observable state variables.

We have demonstrated that, assuming CPMCR is not verified, a na\"{i}ve application of TERC is sufficient to identify the minimal set (Theorem 1). However, this assumption is not universally satisfied. Consequently, we have also presented a solution in case of CPMCR. We have presented an implementation, underpinned by theoretical results, which guarantees the derivation of the minimal set with a weak assumption (Theorem 2). 
In general, TERC represents a novel approach toward feature selection that scales linearly in time without relying on the restricting assumption of having only pairwise dependencies in the set of variables. Furthermore, even if $C_1$ is not satisfied, we still select an information-theoretically optimal set; but we cannot ensure it is also minimal.

Finally, we have implemented our method using a neural estimator and we have then evaluated TERC using synthetic data and across diverse environments representative of various RL problem classes. In particular, we have demonstrated that our proposed method consistently identifies the optimal set of variables in presence of redundant and synergistic relationships. Moreover, we have shown that the resulting compact state representations yield concrete deployment benefits, with inference speedups of up to $2.6\times$ when dimensionality reduction is substantial. These efficiency gains are measured when a policy is retrained or used for inference on the TERC-selected subset; they do not account for the cost of the initial full-state training run used to generate the trajectories on which TERC is applied.


\section*{Acknowledgments} Charles Westphal's PhD studies are supported by the UK Engineering and Physical Sciences Research Council Grant EP/S022503/1.

\bibliography{main}

@article{kraskov2004estimating,
  title   = {Estimating mutual information},
  author  = {Kraskov, Alexander and St{\"o}gbauer, Harald and Grassberger, Peter},
  journal = {Physical Review E},
  volume  = {69},
  number  = {6},
  pages   = {066138},
  year    = {2004}
}

@inproceedings{reda2020learning,
  title={Learning to locomote: Understanding how environment design matters for deep reinforcement learning},
  author={Reda, Daniele and Tao, Tianxin and van de Panne, Michiel},
  booktitle={MIG'20},
  year = {2020}
}

@inproceedings{chen18,
  title={Learning to explain: An information-theoretic perspective on model interpretation},
  author={Chen, Jianbo and Song, Le and Wainwright, Martin and Jordan, Michael},
  booktitle={ICML'18},
  year = {2018}
}

@inproceedings{gao16,
 author = {Gao, Shuyang and Ver Steeg, Greg and Galstyan, Aram},
 booktitle = {NeurIPS'16},
 editor = {D. Lee and M. Sugiyama and U. Luxburg and I. Guyon and R. Garnett},
 title = {Variational Information Maximization for Feature Selection},
 
 year = {2016}
}

@article{peng05,
  title={Feature selection based on mutual information criteria of max-dependency, max-relevance, and min-redundancy},
  author={Peng, Hanchuan and Long, Fuhui and Ding, Chris},
  journal={IEEE Transactions on Pattern Analysis and Machine Intelligence},
  volume={27},
  number={8},
  pages={1226--1238},
  year={2005},
  publisher={IEEE}
}

@inproceedings{ortiz2018learning,
  title={Learning state representations for query optimization with deep reinforcement learning},
  author={Ortiz, Jennifer and Balazinska, Magdalena and Gehrke, Johannes and Keerthi, S. Sathiya},
  booktitle={ DEEM'18},
  year={2018}
}

@article{liu2020state,
  title={State representation modeling for deep reinforcement learning based recommendation},
  pages = {106170},
volume = {205},
  author={Liu, Feng and Tang, Ruiming and Li, Xutao and Zhang, Weinan and Ye, Yunming and Chen, Haokun and Guo, Huifeng and Zhang, Yuzhou and He, Xiuqiang},
  journal={Knowledge-Based Systems},
  
  year={2020},
}

@article{lesort2018state,
  title={State representation learning for control: An overview},
  volume = {108},
pages = {379-392},
  author={Lesort, Timoth{\'e}e and D{\'\i}az-Rodr{\'\i}guez, Natalia and Goudou, Jean-Franois and Filliat, David},
  journal={Neural Networks},
  year={2018},
}

@inproceedings{Schwarzer2020,
  author = {Schwarzer, Max and Anand, Ankesh and Goel, Rishab and Hjelm, R. Devon and Courville, Aaron and Bachman, Philip},
  title = {Data-Efficient Reinforcement Learning with Self-Predictive Representations},
  booktitle = {ICLR'20},
  year = {2020},
}

@inproceedings{Pan2017,
      title={Virtual to Real Reinforcement Learning for Autonomous Driving}, 
      author={Xinlei Pan and Yurong You and Ziyan Wang and Cewu Lu},
      year={2017},
      booktitle={arXiv:1704.03952}
}

@inproceedings{Anastassacos2020,
   
   author = {Nicolas Anastassacos and Stephen Hailes and Mirco Musolesi},
   BOOKTITLE = {AAAI'20},
   title = {Partner selection for the emergence of cooperation in multi-agent systems using reinforcement learning},
   year = {2020},
}

@article{press2012iterated,
  title={{Iterated Prisoner’s Dilemma contains strategies that dominate any evolutionary opponent}},
  volume = {109},
number = {26},
pages = {10409-10413},
  author={Press, William H and Dyson, Freeman J.},
  journal={Proceedings of the National Academy of Sciences},
  year={2012},
}

@inproceedings{schulman2017,
      title={Proximal Policy Optimization Algorithms}, 
      author={John Schulman and Filip Wolski and Prafulla Dhariwal and Alec Radford and Oleg Klimov},
      year={2017},
      booktitle={arXiv:1707.06347}
}

@article{shamir1979share,
  title={How to share a secret},
  volume = {22},
number = {11},
pages = {612–613},
  author={Shamir, Adi},
  journal={Communications of the Association for Computing Machinery},
  year={1979},
}

@inproceedings{blakely1979,
  author={Blakley, George},
  BOOKTITLE={MaRK'79}, 
  title={Safeguarding cryptographic keys}, 
  year={1979}}

@inproceedings{finn2016,
  author={Finn, Chelsea and Xin Yu Tan and Yan Duan and Darrell, Trevor and Levine, Sergey and Abbeel, Pieter},
  booktitle={ICRA'16 }, 
  title={Deep spatial autoencoders for visuomotor learning}, 
  year={2016}}

@inproceedings{stooke2021,
  title = 	 {Decoupling Representation Learning from Reinforcement Learning},
  author =       {Stooke, Adam and Lee, Kimin and Abbeel, Pieter and Laskin, Michael},
  BOOKTITLE = 	 { ICML'21},
  year = 	 {2021},
  
}

@inproceedings{laskin2018,
  title = 	 {{CURL}: Contrastive Unsupervised Representations for Reinforcement Learning},
  author =       {Laskin, Michael and Srinivas, Aravind and Abbeel, Pieter},
  BOOKTITLE = 	 {ICML'20},
  year = 	 {2020}
}

@inproceedings{brockman2016openai,
      title={{OpenAI Gym}}, 
      author={Greg Brockman and Vicki Cheung and Ludwig Pettersson and Jonas Schneider and John Schulman and Jie Tang and Wojciech Zaremba},
      year={2016},
      booktitle={arXiv:1606.01540}
}

@article{Mnih2015,
   author = {Volodymyr Mnih and Koray Kavukcuoglu and David Silver and Andrei A. Rusu and Joel Veness and Marc G. Bellemare and Alex Graves and Martin Riedmiller and Andreas K. Fidjeland and Georg Ostrovski and Stig Petersen and Charles Beattie and Amir Sadik and Ioannis Antonoglou and Helen King and Dharshan Kumaran and Daan Wierstra and Shane Legg and Demis Hassabis},
   journal = {Nature},
   title = {Human-level control through deep reinforcement learning},
   year = {2015},
   
  number={6},
  volume={518},
  pages={529-533}
}

@inproceedings{dean1997,
author = {Dean, Thomas and Givan, Robert},
title = {{Model Minimization in Markov Decision Processes}},
BOOKTITLE={AAAI'97},
year = {1997},
}

@article{
bellman1959,
author = {Richard Bellman  and Robert Kalaba },
title = {A Mathematical Theory of Adaptive Control Processes},
journal = {Proceedings of the National Academy of Sciences},
volume = {45},
number = {8},
pages = {1288-1290},
year = {1959}}

@article{bell1995,
author = {Bell, Anthony and Sejnowski, Terrence},
title = {An Information-Maximization Approach to Blind Separation and Blind Deconvolution},
number = {12},
pages = {1129-59},

volume = {7},
journal = {Neural Computation},
year = {1995}
}

@inproceedings{devlin2018,
  author = {Devlin, Jacob and Chang, Ming-Wei and Lee, Kenton and Toutanova, Kristina},
  title = {BERT: Pre-training of Deep Bidirectional Transformers for Language Understanding},
  
  BOOKTITLE = {ACL'19},
  
  year = {2019},
}

@inproceedings{sun2019,
      title={Learning Video Representations using Contrastive Bidirectional Transformer}, 
      author={Chen Sun and Fabien Baradel and Kevin Murphy and Cordelia Schmid},
      year={2019},
      booktitle={arXiv:1906.05743}
}

@inproceedings{hjelm2018,
  author = {Hjelm, R. Devon and Fedorov, Alex and Lavoie-Marchildon, Samuel and Grewal, Karan and Bachman, Phil and Trischler, Adam and Bengio, Yoshua},
  
  title = {Learning deep representations by mutual information estimation and maximization},
  
  year = {2019},
  
  BOOKTITLE = {ICLR'19},
  
}

@inproceedings{bachman2019,
  title={Learning Representations by Maximizing Mutual Information across Views},
  author={Bachman, Philip and Hjelm, R. Devon and Buchwalter, William},
  BOOKTITLE={NeurIPS'19},
  year={2019}
}

@inproceedings{
Song2020,
title={Understanding the Limitations of Variational Mutual Information Estimators},
author={Jiaming Song and Stefano Ermon},
BOOKTITLE={ICLR'20},
year={2020},
}

@inproceedings{oord2018,
      title={Representation Learning with Contrastive Predictive Coding}, 
      author={Aaron van den Oord and Yazhe Li and Oriol Vinyals},
      year={2019},
booktitle = {arXiv:1807.03748}
}

@inproceedings{belghazi2018,
author = {Belghazi, Ishmael and Rajeswar, Sai and Baratin, Aristide and Hjelm, R. Devon and Courville, Aaron},
year = {2018},
title = {MINE: Mutual Information Neural Estimation}, 
BOOKTITLE = {ICML'18},
}

@inproceedings{Poole2019,
author = {Poole, Ben and Ozair, Sherjil and Oord, Aaron and Alemi, Alexander and Tucker, George},
year = {2019},
BOOKTITLE = {ICML'19},
title = {On Variational Bounds of Mutual Information}
}

@article{bean1987,
author = {James C. Bean, John R. Birge, Robert L. Smith},
year = {1987},
journal = {Operations Research},
title = {Aggregation in Dynamic Programming.},
 number = {2},
 pages = {215-220},
 volume = {35},
}

@article{givan2003,
title = {{Equivalence notions and model minimization in Markov decision processes}},
volume = {147},
number = {1},
pages = {163-223},
journal = {Artificial Intelligence},
year = {2003},
author = {Robert Givan and Thomas Dean and Matthew Greig},

}

@inproceedings{Ravindran2003,
author = {Ravindran, Balaraman and Barto, Andrew G.},
title = {{SMDP Homomorphisms: An Algebraic Approach to Abstraction in Semi-Markov Decision Processes}},
year = {2003},
BOOKTITLE = {IJCAI'03},
}

@phdthesis{mccallum1996,
author = {McCallum, Andrew Kachites and Ballard, Dana},
title = {Reinforcement Learning with Selective Perception and Hidden State},
year = {1996},
school = {The University of Rochester},
}

@inproceedings{jong2005,
author = {Jong, Nicholas K. and Stone, Peter},
title = {State Abstraction Discovery from Irrelevant State Variables},
year = {2005},
BOOKTITLE = {IJCAI'05}}

@inproceedings{li2006,
author = {Li, Lihong and Walsh, Thomas and Littman, Michael},
year = {2006},
title = {{Towards a Unified Theory of State Abstraction for MDPs}},
booktitle = {ISAIM'06}
}

@inproceedings{gelada2019,
author = {Gelada, Carles and Kumar, Saurabh and Buckman, Jacob and Nachum, Ofir and Bellemare, Marc},
year = {2019},
title = {DeepMDP: Learning Continuous Latent Space Models for Representation Learning}, 
BOOKTITLE = {ICML'19 },
}

@inproceedings{grooten2023,
  title={Automatic Noise Filtering with Dynamic Sparse Training in Deep Reinforcement Learning},
  author={Grooten, Bram and Sokar, Ghada and Dohare, Shibhansh and Mocanu, Elena and Taylor, Matthew E and Pechenizkiy, Mykola and Mocanu, Decebal Constantin},
  BOOKTITLE = {AAMAS'23},
  year={2023}
}

@inproceedings{kwon22,
 author = {Kwon, Yongchan and Zou, James Y.},
 booktitle = {NeurIPS'22},
 title = {{WeightedSHAP: analyzing and improving Shapley based feature attributions}},
 year = {2022}
}

@article{Granger1969,
  title={Investigating causal relations by econometric models and cross-spectral methods},
  author={Clive William John Granger},
  year={1969},
  number = {3},
 pages = {424-438},
 volume = {37},
  journal = {Econometrica}
}

@article{neyman_pearson_1933, title={The testing of statistical hypotheses in relation to probabilities a priori}, 
volume={29},
number={4}, 
journal={Mathematical Proceedings of the Cambridge Philosophical Society},  author={Neyman, Jerzy and Pearson, Egon}, 
year={1933},
pages={492–510}}

@ARTICLE{Linsker1988,
  author={Linsker, Ralph},
  journal={Computer}, 
  title={Self-organization in a perceptual network}, 
  pages = {105-117},
volume = {21},
  number = {3},
  year={1988},}

@article{shannon1948,
  title={A mathematical theory of communication},
  author={Shannon, Claude E.},
  journal={{The Bell System Technical Journal}},
  volume={27},
  number={3},
  pages={379--423},
  year={1948},
  publisher={Nokia Bell Labs}
}

@book{sutton2018,
  title={{Reinforcement Learning: An Introduction}},
  author={Sutton, Richard S. and Barto, Andrew G.},
  year={2018},
  publisher={MIT Press},
  address = {Cambridge},
}

@article{schreiber2000,
  title={Measuring information transfer},
  author={Schreiber, Thomas},
   volume = {85},
  number = {2},
  pages = {461-464},
  journal={Physical Review Letters},
  year={2000},
}

@inproceedings{williams2010,
      title={Nonnegative Decomposition of Multivariate Information}, 
      author={Paul L. Williams and Randall D. Beer},
      year={2010},
      booktitle={arXiv:1004.2515},
}

@article{probst2019,
  title={Tunability: Importance of hyperparameters of machine learning algorithms},
  author={Probst, Philipp and Boulesteix, Anne-Laure and Bischl, Bernd},
  journal={Journal of Machine Learning Research},
  volume={20},
  number={1},
  pages={1934--1965},
  year={2019},
  
}

@inproceedings{catav21,
  title = 	 {Marginal Contribution Feature Importance - an Axiomatic Approach for Explaining Data},
  author =       {Catav, Amnon and Fu, Boyang and Zoabi, Yazeed and Meilik, Ahuva Libi Weiss and Shomron, Noam and Ernst, Jason and Sankararaman, Sriram and Gilad-Bachrach, Ran},
  year = 	 {2021},
  booktitle = 	 {ICML'21},
  
}

@inproceedings{janssen2023,
  title={Ultra-marginal Feature Importance: Learning from Data with Causal Guarantees},
  author={Janssen, Joseph and Guan, Vincent and Robeva, Elina},
  BOOKTITLE={AISTATS'23},
  pages={10782-10814},
  year={2023},
}

@article{cohen2007,
author = {Cohen, Shay and Dror, Gideon and Ruppin, Eytan},
year = {2007},
number = {8},
pages = {1939-61},
title = {Feature Selection Via Coalitional Game Theory},
volume = {19},
journal = {Neural Computation}
}

@article{Keinan2004,
title = {{Causal localization of neural function: the Shapley value method}},
journal = {Neurocomputing},
volume = {58-60},
pages = {215-222},
year = {2004},
number = {3},
author = {Alon Keinan and Claus C. Hilgetag and Isaac Meilijson and Eytan Ruppin},

}

@article{Shapley1953,

title = {A Value for n-Person Games},
journal = {Contributions to the Theory of Games},
author = {Shapley, Lloyd},
volume = {2},
pages = {307-318},

year = {1953},
number = {28}
}

@inproceedings{Lunberg2017,
author = {Lundberg, Scott M. and Lee, Su-In},
title = {A Unified Approach to Interpreting Model Predictions},
year = {2017},


BOOKTITLE = {NeurIPS'17},

}

@inproceedings{covert2020,
author = {Covert, Ian C. and Lundberg, Scott and Lee, Su-In},
title = {Understanding Global Feature Contributions with Additive Importance Measures},
year = {2020},

BOOKTITLE = {NeurIPS'20},

}

@inproceedings{Frye2020,
author = {Frye, Christopher and Rowat, Colin and Feige, Ilya},
title = {{Asymmetric Shapley Values: Incorporating Causal Knowledge into Model-Agnostic Explainability}},
year = {2020},

BOOKTITLE= {NeurIPS'20},
}

@inproceedings{Kumar2020,
author = {Kumar, Elizabeth and Venkatasubramanian, Suresh and Scheidegger, Carlos and Friedler, Sorelle A.},
title = {{Problems with Shapley-Value-Based Explanations as Feature Importance Measures}},
year = {2020},

BOOKTITLE = {ICML'20},

}

@inproceedings{chen2020,
      title={True to the Model or True to the Data?}, 
      author={Hugh Chen and Joseph D. Janizek and Scott Lundberg and Su-In Lee},
      year={2020},
      booktitle={arXiv:2006.16234}
}

@article{pedregosa2011,
  title={{Scikit-learn: Machine learning in Python}},
  author={Pedregosa, Fabian and Varoquaux, Ga{\"e}l and Gramfort, Alexandre and Michel, Vincent and Thirion, Bertrand and Grisel, Olivier and Blondel, Mathieu and Prettenhofer, Peter and Weiss, Ron and Dubourg, Vincent and others},
  journal={Journal of Machine Learning Research},
  volume={12},
  pages={2825--2830},
  year={2011},
  publisher={JMLR. org}
}

@Article{Apley2020,
  author={Daniel W. Apley and Jingyu Zhu},
  title={{Visualizing the effects of predictor variables in black box supervised learning models}},
  journal={Journal of the Royal Statistical Society Series B},
  year=2020,
  volume={82},
  number={4},
  pages={1059-1086}
}

@article{breiman2001,
  title={Random forests},
  author={Breiman, Leo},
  journal={Machine Learning},
  volume={45},
  pages={5-32},
  year={2001},
  number={1},
  publisher={Springer}
}

@article{jansen2022,
author = {Janssen, Alexander and Hoogendoorn, Mark and Cnossen, Marjon H. and Mathôt, Ron A. A. and { the OPTI-CLOT Study Group and SYMPHONY Consortium}},
title = {{Application of SHAP values for inferring the optimal functional form of covariates in pharmacokinetic modeling}},
journal = {CPT: Pharmacometrics \& Systems Pharmacology},
volume = {11},
number = {8},
pages = {1100-1110},
year = {2022}
}

@article{debeer20,
author = {Debeer, Dries and Strobl, Carolin},
year = {2020},
month = {07},
pages = {1-30},
title = {Conditional permutation importance revisited},
volume = {21},
journal = {BMC Bioinformatics},
number = "1",
}

@inproceedings{shrikumar2017,
author = {Shrikumar, Avanti and Greenside, Peyton and Kundaje, Anshul},
title = {Learning Important Features through Propagating Activation Differences},
year = {2017},
BOOKTITLE = {ICML'17}

}

@inproceedings{Ribeiro2016,
author = {Ribeiro, Marco Tulio and Singh, Sameer and Guestrin, Carlos},
title = {{``Why Should I Trust You?'': Explaining the Predictions of Any Classifier}},
year = {2016},

BOOKTITLE = {SIGKDD'16},

}

@inproceedings{Sundararajan2020,
  title = 	 {{The Many Shapley Values for Model Explanation}},
  author =       {Sundararajan, Mukund and Najmi, Amir},
  BOOKTITLE = 	 {ICML'20},
  year = 	 {2020},
  
}

@inproceedings{plumb2018,
author = {Plumb, Gregory and Molitor, Denali and Talwalkar, Ameet},
title = {Model Agnostic Supervised Local Explanations},
year = {2018},
BOOKTITLE = {NeurIPS'18},

}

@article{axelrod1981,
  title={The evolution of cooperation},
  author={Axelrod, Robert and Hamilton, William D.},
  journal={Science},
  volume={211},
  number={4489},
  pages={1390--1396},
  year={1981},
  publisher={American Association for the Advancement of Science}
}

@article{brown12,
  author  = {Gavin Brown and Adam Pocock and Ming-Jie Zhao and Mikel Luj{{\'a}}n},
  title   = {Conditional Likelihood Maximisation: A Unifying Framework for Information Theoretic Feature Selection},
  journal = {Journal of Machine Learning Research},
  year    = {2012},
  volume  = {13},
  number  = {2},
  pages   = {27--66},
}

@InProceedings{covert23,
  title = 	 {Learning to Maximize Mutual Information for Dynamic Feature Selection},
  author =       {Covert, Ian Connick and Qiu, Wei and Lu, Mingyu and Kim, Na Yoon and White, Nathan J. and Lee, Su-In},
  booktitle = 	 {ICML'23},
  year = 	 {2023}}

@article{wollstadt23,
  title={A rigorous information-theoretic definition of redundancy and relevancy in feature selection based on (partial) information decomposition},
  author={Wollstadt, Patricia and Schmitt, Sebastian and Wibral, Michael},
  journal={Journal of Machine Learning Research},
  volume={24},
  number={131},
  pages={1--44},
  year={2023}
}

@ARTICLE{battiti94,
  author={Battiti, Roberto},
  journal={IEEE Transactions on Neural Networks}, 
  title={Using mutual information for selecting features in supervised neural net learning}, 
  year={1994},
  volume={5},
  number={4},
  pages={537-550}}

@inproceedings{yang20,
  title={Reinforcement learning in feature space: Matrix bandit, kernels, and regret bound},
  author={Yang, Lin and Wang, Mengdi},
  booktitle={ICML'20},
 year={2020},
}

@ARTICLE{patterson_2022_carbon,
  author={Patterson, David and Gonzalez, Joseph and Hölzle, Urs and Le, Quoc and Liang, Chen and Munguia, Lluis-Miquel and Rothchild, Daniel and So, David R. and Texier, Maud and Dean, Jeff},
  journal={Computer}, 
  title={The Carbon Footprint of Machine Learning Training Will Plateau, Then Shrink}, 
  year={2022},
  volume={55},
  number={7},
  pages={18-28}}

@inproceedings{bonetti2023causal,
  title={Causal Feature Selection via Transfer Entropy},
  author={Bonetti, Paolo and Metelli, Alberto Maria and Restelli, Marcello},
  booktitle={IJCNN'24},
  year={2024}
}

@article{borboudakis2019forward,
  title={Forward-backward selection with early dropping},
  author={Borboudakis, Giorgos and Tsamardinos, Ioannis},
  journal={Journal of Machine Learning Research},
  volume={20},
  number={1},
  pages={276--314},
  year={2019},
}

@article{tsamardinos2019greedy,
  title={{A greedy feature selection algorithm for Big Data of high dimensionality}},
  author={Tsamardinos, Ioannis and Borboudakis, Giorgos and Katsogridakis, Pavlos and Pratikakis, Polyvios and Christophides, Vassilis},
  journal={Machine Learning},
  volume={108},
  pages={149--202},
  year={2019},
  publisher={Springer}
}

@inproceedings{hao2021sparse,
  title={Sparse feature selection makes batch reinforcement learning more sample efficient},
  author={Hao, Botao and Duan, Yaqi and Lattimore, Tor and Szepesv{\'a}ri, Csaba and Wang, Mengdi},
  booktitle={ICML'21},
  year={2021},
}

@article{wookey2015regularized,
  title={Regularized feature selection in reinforcement learning},
  author={Wookey, Dean S. and Konidaris, George D.},
  journal={Machine Learning},
  volume={100},
  number={2-3},
  pages={655--676},
  year={2015},
  publisher={Springer}
}

@article{tesmer2004,
  title={{AMIFS: Adaptive feature selection by using mutual information}},
  author={Tesmer, Mauricio and Est{\'e}vez, Pablo A.},
  journal={IJCNN'04},
  year={2004}
}

@inproceedings{hein2018,
  title={Generating interpretable fuzzy controllers using particle swarm optimization and genetic programming},
  author={Hein, Daniel and Udluft, Steffen and Runkler, Thomas A},
  booktitle={GECCO '18},
  year={2018}
}

@inproceedings{duell2010,
  title={The Markov Decision Process Extraction Network.},
  author={Duell, Siegmund and Hans, Alexander and Udluft, Steffen},
  booktitle={ESANN'10},
  year={2010}
}

@inproceedings{schaefer2007neural,
  title={A neural reinforcement learning approach to gas turbine control},
  author={Sch{\"a}fer, Anton Maximilian and Schneegass, Daniel and Sterzing, Volkmar and Udluft, Steffen},
  booktitle={IJCNN'07},
  year={2007},
}

@inproceedings{schafer2005solving,
  title={Solving partially observable reinforcement learning problems with recurrent neural networks},
  author={Sch{\"a}fer, Anton Maximilian and Udluft, Steffen},
  booktitle={Workshop Proceedings of the European Conference on Machine Learning},
  year={2005}
}

@article{ma2023sequential,
  title={Sequential knockoffs for variable selection in reinforcement learning},
  author={Ma, Tao and Cai, Hengrui and Qi, Zhengling and Shi, Chengchun and Laber, Eric B.},
  journal={arXiv preprint arXiv:2303.14281},
  year={2023}
}
\bibliographystyle{tmlr}
\newpage
\appendix



\section{Terminology and Notation Summary}
\label{appendix:terminology}

\paragraph{Terminology.} Throughout this paper, we distinguish between several closely related but conceptually distinct notions, defined as follows.

\textit{Observable State}: $\mathcal{X} = \{X_1,\dots,X_N\}$ denotes the set of all state variables observable at a given time-step $t$.

\textit{State}: the concrete vector used as input to the policy at time $t$, $s^t \in \mathcal{S}$, built from (a subset of) the observables, possibly stacked over a history window.

\textit{Markovian state}: a state representation such that the transition and reward distributions satisfy the Markov property, i.e.\ conditional on the current state the future is independent of the past.

\textit{Quasi-Markovian state}: a compressed representation (e.g.\ produced by an RNN over a window of observables) that approximates a Markovian state for a partially observable problem~\citep{schafer2005solving, schaefer2007neural}.

\textit{Minimal Markovian state}: a Markovian state of smallest cardinality among those that preserve $H(A\mid \mathcal{X})$ for the trained policy; this is the target representation $\mathcal{X}_*$ derived by TERC.

\paragraph{Notation summary.} Table~\ref{tab:notation} summarizes the main symbols used throughout the paper.

\begin{table}[h]
\centering
\small
\caption{Notation summary.}
\label{tab:notation}
\begin{tabular}{ll}
\toprule
\textbf{Symbol} & \textbf{Meaning} \\
\midrule
$\mathcal{X}=\{X_1,\dots,X_N\}$ & Set of all observable state variables. \\
$X_i$, $x_i$ & Observable random variable and its realization. \\
$A$, $a^t$ & Action random variable and realization at time $t$. \\
$s^t \in \mathcal{S}$ & State vector at time $t$, $s^t=[x^t_1,\dots,x^t_N]$. \\
$\mathcal{P}, \mathcal{P}' \subseteq \mathcal{X}$ & Subsets of observables. \\
$\mathscr{P}(\mathcal{X})$ & Power set of $\mathcal{X}$. \\
$\mathcal{X}_{\setminus \mathcal{P}}$ & Set difference $\mathcal{X}\setminus \mathcal{P}$. \\
$\mathcal{X}_{\setminus(\mathcal{P},\mathcal{P}')}$ & Shorthand for $\mathcal{X}\setminus(\mathcal{P}\cup\mathcal{P}')$. \\
$H(\cdot)$, $H(\cdot\mid\cdot)$ & Shannon (conditional) entropy. \\
$TE_{Y\to X}$ & Transfer entropy from $Y$ to $X$. \\
$\Phi_{X_i;\mathcal{X}\to A}$ & TERC measure for $X_i$ (Eq.~\ref{eqn:TE_measure}). \\
$\Psi(A\mid\mathcal{X})$ & CPMCR condition (Eq.~\ref{eqn:init_pcr}). \\
$C_1$ & Condition~1 (Eq.~\ref{eqn:a1}). \\
$\mathcal{X}_{\Phi}$ & Naïve TERC-selected subset (Eq.~\ref{eqn:origsubset}). \\
$\mathcal{X}_{A_1}$ & Subset returned by Algorithm~\ref{alg:algorithm}. \\
$\mathcal{X}_*$ & Target minimal subset (Eq.~\ref{eqn:problem_statement}). \\
$NM$ & Null-model random variable (Sec.~\ref{subsection:practical}). \\
\bottomrule
\end{tabular}
\end{table}

\section{Empirical Verification of Condition 1}
\label{appendix:condition_1}

In this section, we empirically verify that Condition 1 (Equation~\ref{eqn:a1}) holds across all environments considered in this paper. Recall that Condition 1 states that no two different-sized subsets of the data variables should provide perfectly redundant information about the actions. If this condition is violated, TERC's Algorithm~\ref{alg:algorithm} may fail to identify the minimal state representation.

\subsection{Methodology}

For each environment, we verify Condition 1 as follows. For all disjoint subset pairs $(\mathcal{P}_1, \mathcal{P}_2)$ where $\mathcal{P}_1, \mathcal{P}_2 \subseteq \mathcal{X}$ and $|\mathcal{P}_1| \neq |\mathcal{P}_2|$, we compute the redundancy:
\begin{equation}
R(\mathcal{P}_1; \mathcal{P}_2 \rightarrow A) = I(\mathcal{P}_1; A) + I(\mathcal{P}_2; A) - I(\mathcal{P}_1, \mathcal{P}_2; A),
\end{equation}
where $I(\cdot; A)$ denotes mutual information with the action variable \citep{williams2010}. Condition 1 is violated if $R(\mathcal{P}_1; \mathcal{P}_2 \rightarrow A) = H(A)$ for any subset pair, where $H(A)$ is the entropy of the action distribution.

We collect 10,000 state-action pairs from trained policies for each environment and estimate mutual information using the KSG estimator \citep{kraskov2004estimating}. For environments with discrete state and action spaces (Secret Key Game, IPD), we compute mutual information directly from empirical distributions.

\subsection{Results}

\begin{figure*}[t]
    \centering
    \subfigure(A){\includegraphics[width=7.25cm]{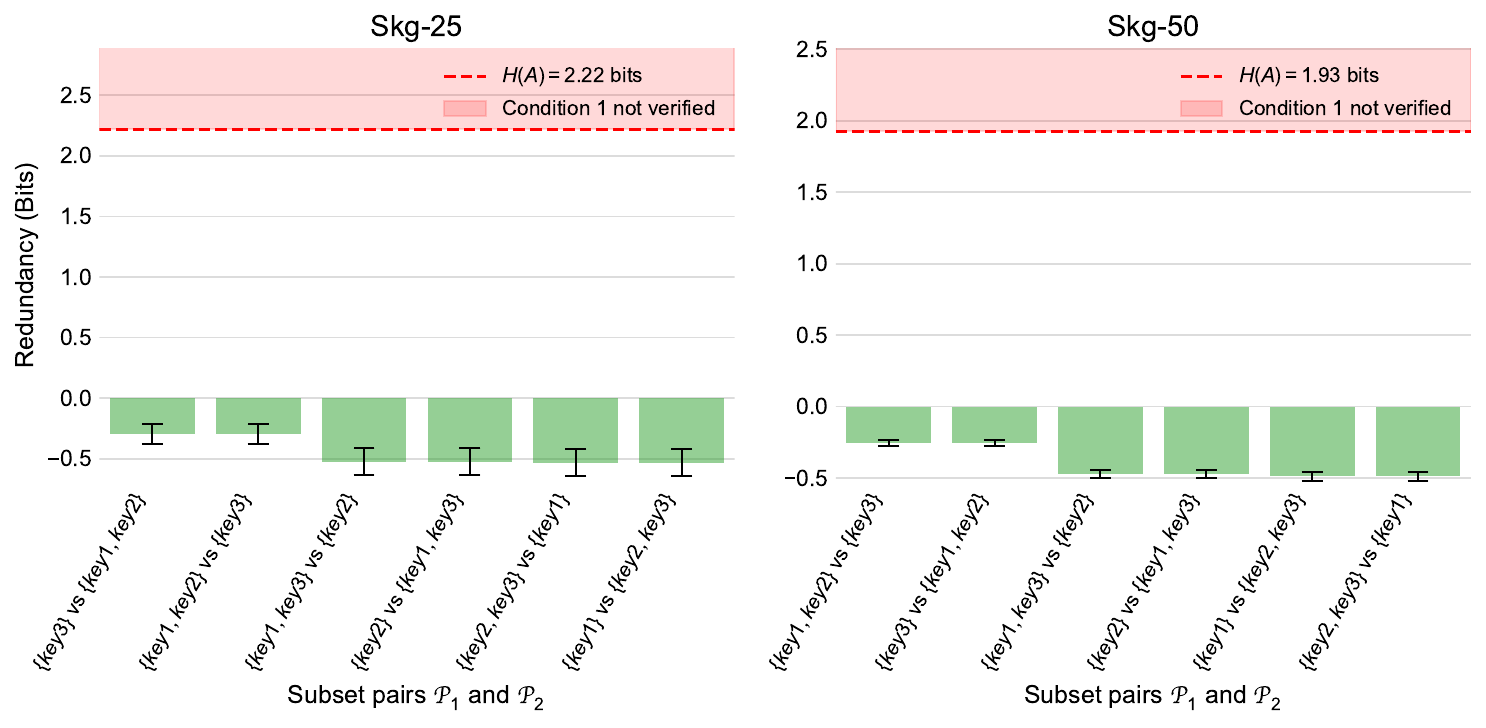}}
    \centering
    \subfigure(B){\includegraphics[width=11.2cm]{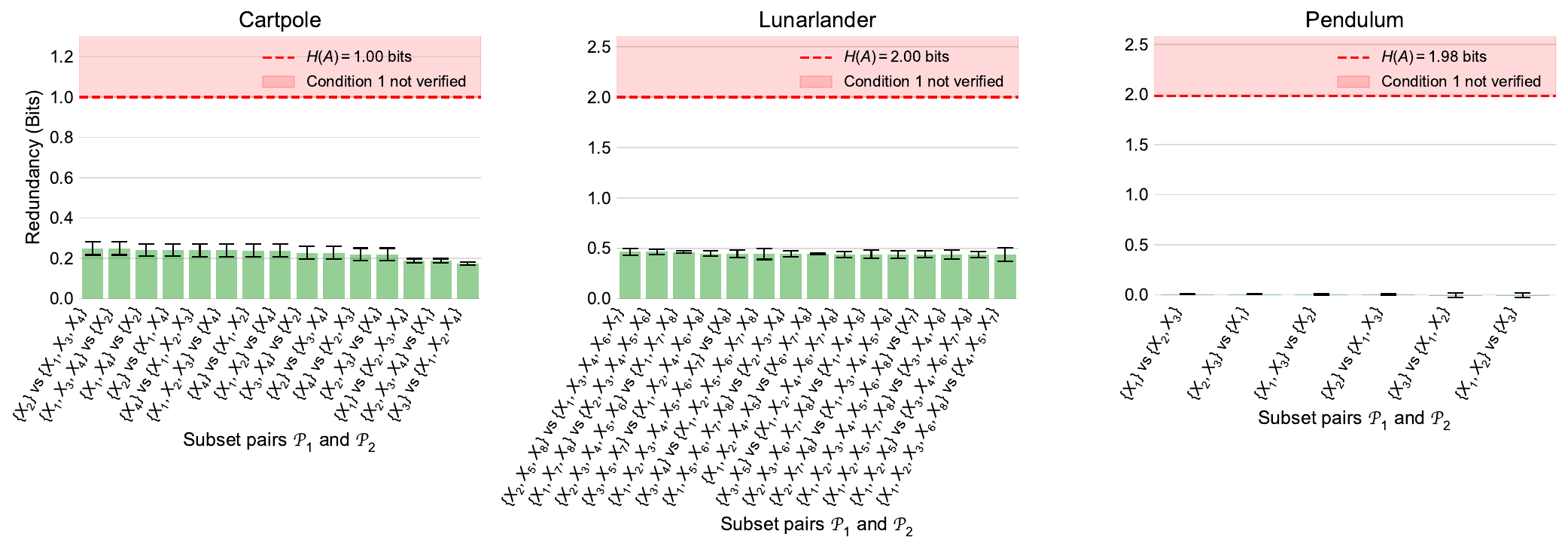}}
    \centering
    \subfigure(C){\includegraphics[width=14.5cm]{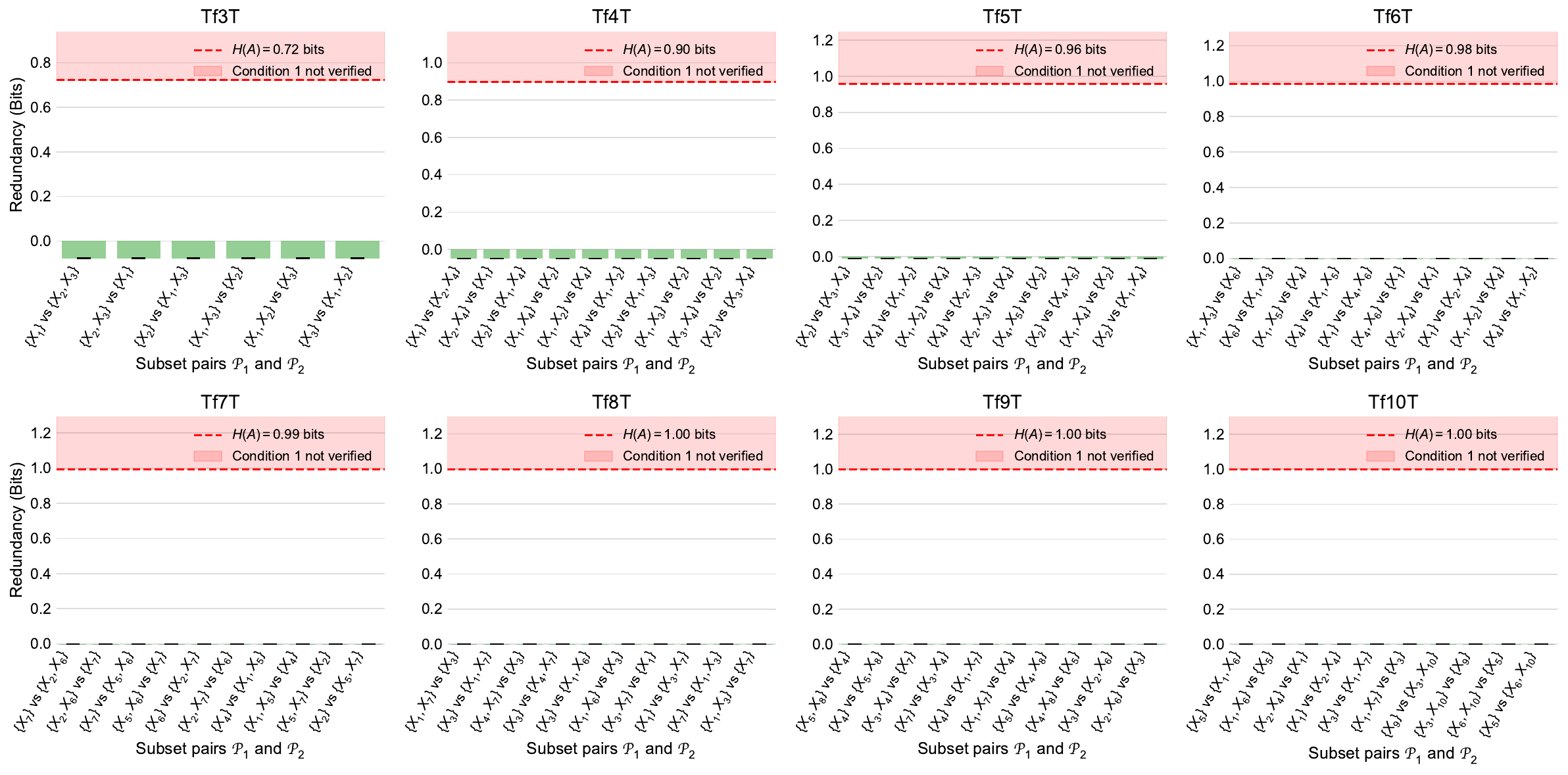}}
    \centering
    \caption{Empirical verification of Condition 1. For each environment, we plot the redundancy $-R(\mathcal{P}_1; \mathcal{P}_2 \rightarrow A)$ for the 10 most redundant subset pairs. The dashed line indicates $H(A)$; bars entering the shaded region would indicate a violation of Condition 1. In accordance with many redundancy synergy indexes, negative values indicate synergy positive values indicate redundancy. No violations are observed across any environment.}
    \label{fig:condition1}
\end{figure*}

Figure~\ref{fig:condition1} presents the redundancy values for the 10 most redundant subset pairs in each environment. The dashed line indicates $H(A)$, the threshold above which Condition 1 would be violated (shaded region). Across all environments,including the Secret Key Game (Figure~\ref{fig:condition1}.A), Gym physics environments (Figure~\ref{fig:condition1}.B), and the Iterated Prisoner's Dilemma against TFNT opponents (Figure~\ref{fig:condition1}.C), no subset pair exceeds this threshold. This confirms that Condition 1 is satisfied in all experimental settings, validating the theoretical guarantees of Algorithm~\ref{alg:algorithm}.

\section{Algorithm for the Estimation of Transfer Entropy Based Measure} \label{appendix:te_alg}
In this section, we outline an algorithm for the estimation of the transfer entropy based measure defined in Equation \ref{eqn:TE_measure}.

\begin{algorithm*}[hbt!]
    \caption{TE measure estimation.}
    \label{alg:te_est}
    \textbf{Input}: Training trajectories $  {\tau} = ({s}^1, {a}^1, {s}^2, {a}^2\ldots {s}^T, {a}^T) = \{{\mathcal{X}}, {A}\}$, where ${a}^1, {a}^2 \in {A}$ and ${s}^1, {s}^2 \in {\mathcal{X}}$. Target variable ${X}_i$.  
    
    \textbf{Output}:  $H({A}|{\mathcal{X}}_{\backslash {X}_i}) - H({A}|{\mathcal{X}})$ 
    \begin{algorithmic}[1] 
    
        \STATE Initialize weights for $\theta$ and $\theta_{\backslash {X}_i}$
        \FOR{$1$ to $N$}
        \STATE Draw mini batch samples of length $b$ from the joint distribution of the actions and the state with all possible variables included $ p_{A,\mathcal{X}} \sim (a^{t_1},x^{t_1}_1,x^{t_1}_i \dots  x^{t_1}_N),\dots,(a^{t_b},x^{t_b}_1,x^{t_b}_i \dots  x^{t_b}_N)$, and repeat for the marginal distribution $p_{A}\otimes p_{\mathcal{X}} \sim (a^{t'_1},x^{t_1}_1,x^{t_1}_i \dots  x^{t_1}_N),\dots,(a^{t'_b},x^{t_b}_1,x^{t_b}_i \dots  x^{t_b}_N)$, where $t'_i \neq t_i$.
        \STATE Draw mini batch samples of length $b$ from the joint distribution of the actions and the state with variable $X_i$ missing.  $ p_{A,\mathcal{X}_{\backslash X_i}} \sim (a^{t_1},x^{t_1}_1 \dots  x^{t_1}_N),\dots,(a^{t_b},x^{t_b}_1 \dots  x^{t_b}_N)$, and repeat for the marginal distribution $p_{A}\otimes p_{\mathcal{X}_{\backslash X_i}} \sim (a^{t'_1},x^{t_1}_1, \dots  x^{t_1}_N),\dots,(a^{t'_b},x^{t_b}_1 \dots  x^{t_b}_N)$, where $t'_i \neq t_i$.
        \STATE $I({A};{\mathcal{X}}) = \frac{1}{b}\sum^b_{j=1}F_{\theta}(a^{t_j},x^{t_j}_1,x^{t_j}_i \dots  x^{t_j}_N)) - \frac{1}{b}\sum^b_{j=1}\log{e^{F_{\theta}((a^{t_j'}_1,x^{t_j}_1,x^{t_j}_i \dots  x^{t_j}_N)}}$
        \STATE $I({A};{\mathcal{X}}_{\backslash {X}_i}) = \frac{1}{b}\sum^b_{j=1}F_{\theta_{\backslash {X}_i}}((a^{t_j},x^{t_j}_1 \dots  x^{t_j}_N)) - \frac{1}{b}\sum^b_{j=1}\log{e^{F_{\theta_{\backslash {X}_i}}(a^{t_j'}_1,x^{t_j}_1 \dots  x^{t_j}_N)}}$
        \ENDFOR
        \STATE \textbf{return} $I({A};{\mathcal{X}}) - I({A};{\mathcal{X}}_{\backslash {X}_i})$
    \end{algorithmic}
\end{algorithm*}

\section{Proof of Non-Negativity of $\Phi_{{X}_i{\mathcal{X}}\rightarrow{A}}$}
\label{appendix:noneg}
In this section, we prove the non-negativity of $\Phi_{{X}_i;{\mathcal{X}} \rightarrow {A}}$. We first write the full expression of the measure as follows: 
\begin{equation}
\label{eqn:noneg1}
\begin{split}
\Phi_{{X}_i;{\mathcal{X}} \rightarrow {A}} =  - \int_{{A} \times {\mathcal{X}}} p_{A,\mathcal{X}}(a^t,x^t_1,x^t_i \dots  x^t_N) \log{\frac{p_{A,\mathcal{X}_{\backslash {X}_i}}(a^t,x^t_1 \dots  x^t_N), p_{\mathcal{X}}( x^t_1,x^t_i \dots  x^t_N)}{p_{A,\mathcal{X}}(a^t,x^t_1,x^t_i \dots  x^t_N), p_{\mathcal{X}_{\backslash {X}_i}}( x^t_1\dots  x^t_N)}} d{A} \times {\mathcal{X}}.
\end{split}
\end{equation}
By applying Jensen's inequality we can then write:
\begin{equation}
\label{eqn:noneg2}
\begin{split}
\Phi_{{X}_i;{\mathcal{X}} \rightarrow {A}} & \geq  - \log \int_{{A} \times {\mathcal{X}}} p_{A,\mathcal{X}}(a^t,x^t_1,x^t_i \dots  x^t_N){\frac{p_{A,\mathcal{X}_{\backslash {X}_i}}(a^t,x^t_1 \dots  x^t_N), p_{\mathcal{X}}( x^t_1,x^t_i \dots  x^t_N)}{p_{A,\mathcal{X}}(a^t,x^t_1,x^t_i \dots  x^t_N), p_{\mathcal{X}_{\backslash {X}_i}}( x^t_1\dots  x^t_N)}} d{A} \times {\mathcal{X}}\\ 
& \geq  - \log \int_{{A} \times {\mathcal{X}}}  {\frac{p_{A,\mathcal{X}_{\backslash {X}_i}}(a^t,x^t_1 \dots  x^t_N), p_{\mathcal{X}}( x^t_1,x^t_i \dots  x^t_N)}{p_{\mathcal{X}_{\backslash {X}_i}}( x^t_1\dots  x^t_N)}} d{A} \times {\mathcal{X}}\\ & \geq  - \log(1) \\ & \geq 0,
\end{split}
\end{equation}
therefore proving the non-negativity of $\Phi_{{X}_i;{\mathcal{X}} \rightarrow {A}}$. $\square$

\section{Proof of Lemma 1} \label{appendix:lem1pmcr}
The steps of this proof can be summarized as follows. Firstly, we show that a single element of a subset for which we observe CPMCR with another variable or subset of variables will not be included in the set ${\mathcal{X}}_{\Phi}$. Then, we will demonstrate that this applies to all the elements of both subsets. Finally, we use the definition of CPMCR to prove Lemma 1.

Let there be a case of CPMCR between two subsets $\mathcal{P},\mathcal{P}' \subseteq {\mathcal{X}}$ such that:
\begin{equation}
\label{eqn:pmcr2}
\begin{split}
 H({A}|{\mathcal{X}}_{\backslash ( \mathcal{P}, \mathcal{P}')} \cup \mathcal{P} \cup \mathcal{P}') & =  H({A}|{\mathcal{X}}_{\backslash ( \mathcal{P}, \mathcal{P}')} \cup \mathcal{P}) \\ & = H({A}|{\mathcal{X}}_{\backslash ( \mathcal{P}, \mathcal{P}')} \cup \mathcal{P}') \\ & <  H({A}|{\mathcal{X}}_{\backslash ( \mathcal{P}, \mathcal{P}')} ).
\end{split}
\end{equation}
Let $\mathcal{P}_{\backslash \mathcal{P}} \in \mathscr{P}({\mathcal{X}}_{\backslash \mathcal{P}})$ and  $\mathcal{P}_{\backslash \mathcal{P}'} \in \mathscr{P}({\mathcal{X}}_{\backslash \mathcal{P}'})$. Given Equation \ref{eqn:elems_pcr_cond} and Equation \ref{eqn:pmcr2}, it must be true that:
\begin{equation}
\label{eqn:pmcr3}
\begin{split}
H({A}|\mathcal{P}_{\backslash \mathcal{P}'}  \cup \mathcal{P}') & \leq H({A}|\mathcal{P}_{\backslash \mathcal{P}'}  \cup \{P'\}) ,
\end{split}
\end{equation}
where $P' \in \mathcal{P}'$ . Combining Equations \ref{eqn:pmcr2} and \ref{eqn:pmcr3} leads us too:
\begin{equation}
\label{eqn:pmcr4}
\begin{split}
H({A}|\mathcal{P}_{\backslash \mathcal{P}} \cup \mathcal{P}) & \leq H({A}|\mathcal{P}_{\backslash \mathcal{P}} \cup \{P'\}),
\end{split}
\end{equation}
we now substitute in ${\mathcal{X}}_{\backslash (\mathcal{P}, P')} = \mathcal{P}_{\backslash \mathcal{P}}$, leading to:
\begin{equation}
\label{eqn:pmcr4'}
\begin{split}
H({A}|{\mathcal{X}}_{\backslash (\mathcal{P}, P')} \cup \mathcal{P}) & \leq H({A}|{\mathcal{X}}_{\backslash (\mathcal{P}, P')} \cup \{P'\})  \\
H({A}|{\mathcal{X}}_{\backslash  P'} ) & \leq H({A}|{\mathcal{X}}_{\backslash \mathcal{P}} ) \\
 & \leq H({A}|{\mathcal{X}}),
\end{split}
\end{equation}
and therefore: 
\begin{equation}
\label{eqn:pmcr5}
\begin{split}
H({A}|{\mathcal{X}}_{\backslash P'} ) - H({A}|{\mathcal{X}}) \leq 0.
\end{split}
\end{equation}
Due to the non-negativity proof presented in Section \ref{appendix:noneg}, we obtain $\Phi_{P';{\mathcal{X}} \rightarrow {A}} = 0$, and consequently element $P'$ is not included in the set ${\mathcal{X}}_{\Phi}$.  This holds $\forall P \in \mathcal{P}$ and $\forall P' \in \mathcal{P}'$. Consequently, neither of the elements of the subset $\mathcal{P}$ or of the subset $\mathcal{P}'$ will be included in the set ${\mathcal{X}}_{\Phi}$. Therefore, ${\mathcal{X}}_{\Phi} = {\mathcal{X}}_{\backslash ( \mathcal{P}, \mathcal{P}')}$, but, through Equation \ref{eqn:pmcr2}, we obtain $H({A}|{\mathcal{X}}_{\Phi} ) > H({A}|{\mathcal{X}} )$. $\square$

\section{Proof of Theorem 1} \label{appendix:t1}

We outline this proof in three steps, which we present as lemmas. The first of these lemmas characterizes the variables that are not to be included in ${\mathcal{X}}_{\Phi}$. By means of the other two, we demonstrate that by not adding these variables to the set ${\mathcal{X}}_{\Phi}$, we still satisfy $H({A}|{\mathcal{X}})=H({A}|{\mathcal{X}}_{\Phi})$.

\noindent \textbf{Lemma 2.} \textit{Let us assume that there exists variables ${X}_i$ in $\mathcal{X}$ that transfers no entropy to the actions $A$ (satisfying $\Phi_{{X}_i;\mathcal{X} \rightarrow A} = 0$).
}

\begin{itemize}
    \item \textit{These variables satisfy one of either:}
    \begin{equation}
    \begin{split}
    H(A|\mathcal{X})  = H(A|\mathcal{X}_{\backslash X_i}) < H(A|X_i)  < H(A)),
    \label{eqn:info_included}
    \end{split}
    \end{equation}
\textit{or}
    \begin{equation}
    \begin{split}
    H({A}|\{{X}_i\} \cup \mathcal{P}_{\backslash {X}_i}) = H({A}|\mathcal{P}_{\backslash {X}_i}));
    \label{eqn:no_info}
    \end{split}
    \end{equation}
\item \textit{assuming no cases of CPMCR ($\neg \Psi(A|\mathcal{X}$)), the following also holds:} 
\begin{equation}
\begin{split}
\Phi_{{X}_i;\mathcal{X} \rightarrow A} = 0   \quad \& \quad \neg \Psi(A|\mathcal{X}) \Leftrightarrow  & H(A|\mathcal{X})  = H(A|\mathcal{X}_{\backslash X_i}) < H(A|X_i)  < H(A ) \quad \text{or} \quad  \\ & H({A}|\{{X}_i\} \cup \mathcal{P}_{\backslash {X}_i}) = H({A}|\mathcal{P}_{\backslash {X}_i}). 
\label{eqn:no_infonew}
\end{split}
\end{equation}
\end{itemize}

\noindent \textit{Proof.}
Let us assume a case of CPMCR such that $\Psi_{X_i, \mathcal{P}'}(A|\mathcal{X})$, as this naturally leads to  $\Phi_{{X}_i;\mathcal{X} \rightarrow A} = 0$.
We now show that to satisfy the condition $\neg \Psi_{X_i, \mathcal{P}'}(A|\mathcal{X})$, the variable $X_i$ must satisfy Equation \ref{eqn:info_included} or Equation \ref{eqn:no_info}.

We first note that we can re-write  $\Phi_{{X}_i;\mathcal{X} \rightarrow A} = 0$ as $H(A|\mathcal{X}_{\backslash X_i}) = H(A|\mathcal{X})$ and therefore the condition $\Phi_{{X}_i;\mathcal{X} \rightarrow A} = 0$ is already satisfied in the definition of CPMCR. Consequently, it is possible to write:
\begin{equation}
\label{eqn:init_pcrl2}
\begin{split}
  \Phi&_{{X}_i;\mathcal{X}  \rightarrow A} = 0   \quad \& \quad \Psi_{X_i, \mathcal{P}'}(A|\mathcal{X})  \\ \Leftrightarrow &   (X_i \in \mathcal{X}, \exists \mathcal{P}'\in \mathscr{P}(\mathcal{X}_{\backslash X_i})  :  H(A|\mathcal{X}) = H(A|\mathcal{X}_{\backslash  \mathcal{P}'}) = H(A|\mathcal{X}_{\backslash X_i})  < H(A|\mathcal{X}_{\backslash ( X_i, \mathcal{P}')} ) \quad  \& \quad \\ & \psi_{\mathcal{P}'}).
\end{split}
\end{equation}
 Suppose that we violate the condition $\psi_\mathcal{P}'$ by adding a non-informative variable $V_{rand}$ to the set $\mathcal{P}'$. Although, this would lead to $\neg \Psi_{X_i, \mathcal{P}'}(A|\mathcal{X})$, it would still be true that $ \Psi_{X_i, \mathcal{P}'_{\backslash V_{rand}}}(A|\mathcal{X})$. Consequently, there will always exist $\mathcal{P}'\in \mathscr{P}$ such that $\psi_\mathcal{P}'$ is satisfied. Given there always exists a scenario in which $\psi_\mathcal{P}'$ is true, it cannot be used to induce the condition $\neg \Psi_{X_i, \mathcal{P}'}(A|\mathcal{X})$. 
 Consequently, for clarity, we remove it from the presentation and we write:
\begin{equation}
\label{eqn:init_pcrl21}
\begin{split}
& \Phi_{{X}_i;\mathcal{X} \rightarrow A} = 0   \quad \& \quad \Psi_{X_i, \mathcal{P}'}(A|\mathcal{X})  \\ & \Leftrightarrow  (X_i \in \mathcal{X}, \exists \mathcal{P}'\in \mathscr{P}(\mathcal{X}_{\backslash X_i})  :  H(A|\mathcal{X}) = H(A|\mathcal{X}_{\backslash  \mathcal{P}'}) = H(A|\mathcal{X}_{\backslash X_i})  < H(A|\mathcal{X}_{\backslash ( X_i, \mathcal{P}')} ) ).
\end{split}
\end{equation}
The next step in this proof is to rewrite Equation \ref{eqn:init_pcrl21} so that it satisfies $\neg \Psi_{X_i, \mathcal{P}'}(A|\mathcal{X})$. 
This will be the case if the equality that relates $H(A|\mathcal{X}_{\backslash  \mathcal{P}'}) = H(A|\mathcal{X}_{\backslash X_i})$ or $H(A|\mathcal{X}_{\backslash X_i})  < H(A|\mathcal{X}_{\backslash ( X_i, \mathcal{P}')} )$  no longer holds. These are the only ways to derive $\neg \Psi_{X_i, \mathcal{P}'}(A|\mathcal{X})$ without leading to $\neg \Phi_{{X}_i;\mathcal{X} \rightarrow A}$. 

 We now assume the equality that relates $H(A|\mathcal{X}_{\backslash  \mathcal{P}'}) = H(A|\mathcal{X}_{\backslash X_i})$ no longer holds.
 We obtain a variable $X_i$ and subset $\mathcal{P}'$ that do not provide the same information about the target variable $H(A|\mathcal{X}_{\backslash  \mathcal{P}'}) > H(A|\mathcal{X}_{\backslash X_i})$. Substituting this into Equation \ref{eqn:init_pcrl21} we have: 
\begin{equation}
\label{eqn:init_pcrl22eqn}
\begin{split}
& \Phi_{{X}_i;\mathcal{X} \rightarrow A} = 0   \quad \& \quad \neg \Psi_{X_i, \mathcal{P}'}(A|\mathcal{X})  \\ & \Leftarrow  (X_i \in \mathcal{X}, \exists \mathcal{P}'\in \mathscr{P}(\mathcal{X}_{\backslash X_i})  :  H(A|\mathcal{X})  = H(A|\mathcal{X}_{\backslash X_i}) < H(A|\mathcal{X}_{\backslash  \mathcal{P}'})  < H(A|\mathcal{X}_{\backslash ( X_i, \mathcal{P}')} ) ),
\end{split}
\end{equation}
Suppose we rewrite Equation \ref{eqn:init_pcrl22eqn} letting $\mathcal{P}' = \mathcal{X}_{\backslash X_i}$. We obtain:
\begin{equation}
\label{eqn:init_pcrl231}
\begin{split}
& \Phi_{{X}_i;\mathcal{X} \rightarrow A} = 0   \quad \& \quad \neg \Psi_{X_i, \mathcal{P}'}(A|\mathcal{X}) \\ & \Leftarrow   ( H(A|\mathcal{X})  = H(A|\mathcal{X}_{\backslash X_i}) < H(A|\mathcal{X}_{\backslash  \mathcal{X}_{\backslash X_i}})  \leq H(A|\mathcal{X}_{\backslash \{ X_i, \mathcal{X}_{\backslash X_i}\}} ) ) \\ & \Leftarrow   ( H(A|\mathcal{X})  = H(A|\mathcal{X}_{\backslash X_i}) < H(A|X_i)  < H(A ) ).
\end{split}
\end{equation}
Hence, the variables that satisfy Equation \ref{eqn:no_info} also satisfy $\Phi_{{X}_i;\mathcal{X} \rightarrow A} = 0   \quad \& \quad \neg \Psi_{X_i, \mathcal{P}'}(A|\mathcal{X})$. 
Now, let us suppose that the inequality $H(A|\mathcal{X}_{\backslash X_i})  < H(A|\mathcal{X}_{\backslash ( X_i, \mathcal{P}')} )$ no longer holds. Specifically, we assume neither the subset $\mathcal{P}'$ or variable $X_i$ provide any information about the target variable. In this case it is true that $H(A|\mathcal{X}_{\backslash X_i})  = H(A|\mathcal{X}_{\backslash ( X_i, \mathcal{P}')} )$ and we can write the following:
\begin{equation}
\label{eqn:init_pcrl23}
\begin{split}
&  \Phi_{{X}_i;\mathcal{X} \rightarrow A} = 0  \quad \& \quad \neg \Psi_{X_i, \mathcal{P}'}(A|\mathcal{X}) \\ & \Leftarrow  (X_i \in \mathcal{X}, \exists \mathcal{P}'\in \mathscr{P}(\mathcal{X}_{\backslash X_i})  :  H(A|\mathcal{X})  = H(A|\mathcal{X}_{\backslash X_i}) = H(A|\mathcal{X}_{\backslash  \mathcal{P}'})  = H(A|\mathcal{X}_{\backslash ( X_i, \mathcal{P}')} ) ).
\end{split}
\end{equation}
We now substitute in $\mathcal{P}_{\backslash (X_i, \mathcal{P}')} = \mathcal{X}_{\backslash (X_i, \mathcal{P}')}$ in Equation \ref{eqn:init_pcrl23} such that:
\begin{equation}
\label{eqn:init_pcrl232}
\begin{split}
&  \Phi_{{X}_i;\mathcal{X} \rightarrow A} = 0  \quad \& \quad \neg \Psi_{X_i, \mathcal{P}'}(A|\mathcal{X}) \\ & \Leftarrow  (X_i \in \mathcal{X}, \exists \mathcal{P}'\in \mathscr{P}(\mathcal{X}_{\backslash X_i})  :  H(A|\mathcal{X})  = H(A|\mathcal{P}_{\backslash (X_i, \mathcal{P}')} \cup \mathcal{P}') = H(A|\mathcal{P}_{\backslash (X_i, \mathcal{P}')} \cup X_i)  = H(A|\mathcal{P}_{\backslash (X_i, \mathcal{P}')} ).
\end{split}
\end{equation}
Given $\mathcal{P}' \in \mathscr{P}(\mathcal{X}_{\backslash X_i})$ and $\mathcal{P}_{\backslash (X_i, \mathcal{P}')} \in \mathscr{P}(\mathcal{X}_{\backslash (X_i, \mathcal{P}')})$, it must be so that $\mathcal{P}_{\backslash (X_i, \mathcal{P}')} \cup \mathcal{P}' \in \mathscr{P}(\mathcal{X}_{\backslash X_i})$. Consequently, we replace $\mathcal{P}_{\backslash (X_i, \mathcal{P}')} \cup \mathcal{P}'$ with $\mathcal{P}_{\backslash X_i}$ in Equation \ref{eqn:init_pcrl232} to give
\begin{equation}
\label{eqn:init_pcrl233}
\begin{split}
\Phi_{{X}_i;\mathcal{X} \rightarrow A} = 0  \quad \& \quad \neg & \Psi_{X_i, \mathcal{P}'}(A|\mathcal{X}) \Leftarrow  (X_i \in \mathcal{X}, \exists \mathcal{P}'\in \mathscr{P}(\mathcal{X}_{\backslash X_i})  : \\ & H(A|\mathcal{P}_{\backslash X_i} \cup X_i)  = H(A|\mathcal{P}_{\backslash X_i}) = H(A|\mathcal{P}_{\backslash (X_i, \mathcal{P}')} \cup X_i)  = H(A|\mathcal{P}_{\backslash (X_i, \mathcal{P}')})).
\end{split}
\end{equation}
Since $\mathcal{P}_{\backslash (X_i, \mathcal{P}')} \subseteq \mathcal{P}_{\backslash X_i}$, we do not include the last two equivalences as they are sub-conditions of the first equivalence. Therefore, we can write
\begin{equation}
\label{eqn:init_pcrl234}
\begin{split}
  \Phi_{{X}_i;\mathcal{X} \rightarrow A} = 0  \quad \& \quad \neg \Psi_{X_i, \mathcal{P}'}(A|\mathcal{X})  \Leftarrow    H(A|\mathcal{P}_{\backslash X_i} \cup X_i)  = H(A|\mathcal{P}_{\backslash X_i}) ).
\end{split}
\end{equation}
Combining the statements in Equations \ref{eqn:init_pcrl234} and \ref{eqn:init_pcrl231} completes the proof of Lemma 2. $\square$



According to Lemma 2, variables that satisfy Equation \ref{eqn:no_info} are not included in the set ${\mathcal{X}}_{\Phi}$, despite this, we now prove that $H({A}|{\mathcal{X}}) = H({A}|{\mathcal{X}}_{\Phi})$ still holds. 

\noindent \textbf{Lemma 3.} \textit{Let us assume there exists a non-empty subset of variables in $\mathcal{X}$ that satisfies Equation \ref{eqn:no_info}. The following holds: $H({A}|{\mathcal{X}}) = H({A}|{\mathcal{X}}_{\Phi})$.}

\noindent \textit{Proof.} To prove Lemma 3, we let variables ${X}_j$ and ${X}_k$ satisfy the property expressed by Equation \ref{eqn:no_info}, and show that despite removing them from ${\mathcal{X}}$ to form ${\mathcal{X}}_{\Phi}$, $H({A}|{\mathcal{X}}) = H({A}|{\mathcal{X}}_{\Phi})$ is still satisfied. We then demonstrate that this result generalizes to cases with more than two variables.

In accordance with Lemma 2, if we remove ${X}_j$ from ${\mathcal{X}}$ we have $H({A}|{\mathcal{X}}) = H({A}|{\mathcal{X}}_{\backslash {X}_j})$. Suppose that we now repeat this process, but instead, we remove ${X}_k$ from ${\mathcal{X}}_{\backslash {X}_j}$, as this will obtain ${\mathcal{X}}_{\Phi}$. To begin, we let $\mathcal{P}_{\backslash {X}_k} \in  \mathscr{P}({\mathcal{X}}_{\backslash {X}_k})$, such that:
\begin{equation}
\begin{split}
 H({A}|\mathcal{P}_{\backslash {X}_k}) & = H({A}|\{{X}_k\} \cup \mathcal{P}_{\backslash {X}_k}).
\end{split}
\end{equation}
We now substitute it in $\mathcal{P}_{\backslash {X}_k} = {\mathcal{X}}_{\backslash \{ X_k, X_j \}}$:
\begin{equation}
\begin{split}
H({A}|{\mathcal{X}}_{\backslash \{ X_k, X_j \}}) & =  H({A}|\{{X}_k\} \cup {\mathcal{X}}_{\backslash \{ X_k, X_j \}} ),
\end{split}
\end{equation}
because $\{{X}_k\} \cup {\mathcal{X}}_{\backslash \{ X_k, X_j \}} = {\mathcal{X}}_{\backslash {X}_j}$ we have:
\begin{equation}
\begin{split}
H({A}|{\mathcal{X}}_{\backslash \{ X_k, X_j \}}) &  = H({A}|{\mathcal{X}}_{\backslash {X}_j} )   \\ &  = H({A}|{\mathcal{X}} ).
\end{split}
\end{equation}
However, given that only the variables ${X}_j$ and ${X}_k$ satisfy the property expressed by Equation \ref{eqn:no_info}, we have ${\mathcal{X}}_{\backslash \{ X_k, X_j \}} = {\mathcal{X}}_{\Phi}$ and therefore $H({A}|{\mathcal{X}}_{\Phi}) = H({A}|{\mathcal{X}} )$. Consequently, we have proved Lemma 3 in the two variable case. 

We now generalize this proof beyond the two variable case. Let variable ${X}_l$ also satisfy the property outlined in Equation \ref{eqn:no_info}, we now have ${\mathcal{X}}_{\backslash \{ X_k, X_j , X_l \}} = {\mathcal{X}}_{\Phi}$. We complete this proof, by first using the property outlined in Equation \ref{eqn:no_info} to show $H({A}| {\mathcal{X}}_{\backslash \{ X_k, X_j \}} ) = H({A}| {\mathcal{X}}_{\backslash \{ X_k, X_j , X_l \}} )$, before then combining this with the proof for the two variable case to show $H({A}|{\mathcal{X}}_{\Phi}) = H({A}|{\mathcal{X}} )$. We can repeatedly apply this proof up to the $N$ variable case. $\square$

According to Lemma 2, the variables that satisfy Equation \ref{eqn:info_included} are not included in the set ${\mathcal{X}}_{\Phi}$.  Despite this, provided no cases of CPMCR exist in the set $\mathcal{X}$, it is possible to prove that $H({A}|{\mathcal{X}}) = H({A}|{\mathcal{X}}_{\Phi})$ still holds. More formally, we demonstrate the following lemma:

\noindent \textbf{Lemma 4.} \textit{Let us assume there exists a non-empty subset of variables in $\mathcal{X}$ that satisfies Equation \ref{eqn:info_included}. Provided no cases of CPMCR exist in $\mathcal{X}$, the following holds: $H({A}|{\mathcal{X}}) = H({A}|{\mathcal{X}}_{\Phi})$.}

\noindent \textit{Proof.} We let variables ${X}_j$ and ${X}_k$ satisfy the property outlined in Equation \ref{eqn:info_included}, and show that if removing them to form ${\mathcal{X}}_{\Phi}$ leads to $H({A}|{\mathcal{X}}) < H({A}|{\mathcal{X}}_{\Phi})$, a contradiction arises. Therefore, it must be true that $H({A}|{\mathcal{X}}) = H({A}|{\mathcal{X}}_{\Phi})$. Subsequently, we generalize to the $N$ variable case. 

If Equation \ref{eqn:info_included} holds for variables $X_j$ and $X_k$, so does $H({A}|{\mathcal{X}})=H({A}|{\mathcal{X}}_{\backslash {X}_j})=H({A}|{\mathcal{X}}_{\backslash {X}_k})$. Now we assume that Lemma 4 is not true. In this case, we have: $H({A}|{\mathcal{X}}) < H({A}|{\mathcal{X}}_{\Phi})$. Currently, we are only interested in the two-variable case, so this becomes $H({A}|{\mathcal{X}}) < H({A}|\mathcal{X}_{\backslash \{X_j, X_k \}})$. Consequently, we have:
\begin{equation}
\begin{split}
H({A}|{\mathcal{X}})=H({A}|{\mathcal{X}}_{\backslash  X_j } )=H({A}|{\mathcal{X}}_{\backslash {X}_k } ) < H({A}|{\mathcal{X}}_{\backslash \{ X_j, {X}_k \}}).
\end{split}
\label{eqn:max_ent52}
\end{equation}

Equation \ref{eqn:max_ent52} reveals that in order to verify the condition $H({A}|{\mathcal{X}}) < H({A}|{\mathcal{X}}_{\Phi})$, there must be CPMCR, such that $\Psi_{X_j, X_k}(A|\mathcal{X})$ ($\psi$ here is invalid as we are concerned with a single variable). Consequently, a contradiction has arisen; therefore, we have proven Lemma 4 in the two-variable case. 

We now generalize this result to the $N$ variable case. Let there be a third variable, $X_l$, that also satisfies the property expressed by Equation \ref{eqn:info_included}. In this case, it must be true that: $H({A}|{\mathcal{X}})=H({A}|{\mathcal{X}}_{\backslash {X}_j})=H({A}|{\mathcal{X}}_{\backslash {X}_k}) = H({A}|{\mathcal{X}}_{\backslash {X}_l})$. Now we reapply an identical method to demonstrate that, if $H({A}|{\mathcal{X}}) < H({A}|{\mathcal{X}}_{\Phi}) = H({A}|\mathcal{X}_{\backslash \{X_j, X_k, X_l \}})$, we must have a case of CPMCR, such that $\Psi_{X_j, X_k, X_l}(A|\mathcal{X})$. Again, a contradiction has arisen, and this proves the lemma in the case of $N$ variables. $\square$

\noindent \textbf{Theorem 1.} \textit{Please refer to Section \ref{sec:prac_app}.}

\noindent \textit{Proof.}
In Lemma 2, we have shown that variables that provide no information about the actions (satisfying Equation \ref{eqn:no_info}) and variables that provide redundant information about the actions (satisfying Equation \ref{eqn:info_included}) will not be included in the set $\mathcal{X}_{\Phi}$. It is therefore trivial to see that the remaining variables in the set $\mathcal{X}_{\Phi}$ provide non-redundant information about the actions, and, therefore, satisfy $H({A}|\mathcal{X}_{\Phi \backslash {X}_i}) > H({A}|\mathcal{X}_{\Phi})$. This means that it is not possible to remove any more variables from $\mathcal{X}_{\Phi}$ without increasing the value of $H({A}|\mathcal{X}_{\Phi})$. Consequently, the set  $\mathcal{X}_{\Phi}$ satisfies the condition $|\mathcal{X}_{\Phi}| = \min\limits_{H({A}|\mathcal{P}) = H({A}|\mathcal{X}_{\Phi})} |\mathcal{P}|$, where $\mathcal{P} \in \mathscr{P}({\mathcal{X}})$. Combining this information with the results of Lemmas 3 and 4, in which we have shown that despite removing these variables the expression $H({A}|{\mathcal{X}}) = H({A}|{\mathcal{X}}_{\Phi})$ still holds, completes the proof of Theorem 1. $\square$

\section{Proof of Theorem 2} \label{appendix:alg_correct}
To carry out this proof we first present Lemma 5, which shows that Algorithm \ref{alg:algorithm} effectively deals with cases of CPMCR between subsets of equal cardinality.
Using this result alongside Theorem 1, we demonstrate that the proof is valid under Condition 1.

\noindent \textbf{Lemma 5.} \textit{Let $\mathcal{P} \in \mathscr{P}({\mathcal{X}}_{A_1})$, where ${\mathcal{X}}_{A_1}$ is generated using Algorithm \ref{alg:algorithm}.
The following holds\footnote{In this lemma, we adopt the definition of $\Psi(A|\mathcal{X}) $ provided in Equation \ref{eqn:init_pcr}.}:
%
}
\begin{equation}
\label{eqn:lem8}
\begin{split}
\forall \mathcal{P} \in \mathscr{P}({\mathcal{X}}_{A_1}) \nexists \mathcal{P}' \in \mathscr{P}({\mathcal{X}}_{A_1}) : \Psi(A|\mathcal{X}) \quad \& \quad |\mathcal{P}| = |\mathcal{P}'| \quad \& \quad  \mathcal{P} \neq \mathcal{P}'.
\end{split}
\end{equation}

\noindent \textit{Proof.} We prove Lemma 5 by demonstrating that Algorithm \ref{alg:algorithm} is able to resolve cases of CPMCR between subsets with equal cardinality. We consider two subsets $\mathcal{P}$ and $\mathcal{P}'$ with equal cardinality ($|\mathcal{P}|=|\mathcal{P}'|$) and CPMCR between them ($\Psi_{\mathcal{P},\mathcal{P}'}(A|\mathcal{X})$). We can write the set ${\mathcal{X}}$ that contains these subsets as:  ${\mathcal{X}} = \{P_1, P'_1, P_2, P'_2\ldots P_N, P'_N\} \cup {\mathcal{X}}_{\backslash ( \mathcal{P}, \mathcal{P}')}$, where $P_i \in \mathcal{P}$ and $P'_i \in \mathcal{P}'$. We can now apply Algorithm \ref{alg:algorithm}: we begin by calculating $\Phi_{P_1; {\mathcal{X}} \rightarrow {A}} = 0$, in accordance with Lemma 1. Consequently, $P_1$ gets removed from the set ${\mathcal{X}}$, so we can write ${\mathcal{X}} = {\mathcal{X}}^0_{\backslash P_1}$, where here  ${\mathcal{X}}^0$ indicates ${\mathcal{X}}$ prior to any algorithmic operations. 
We now calculate $\Phi_{P'_1; {\mathcal{X}} \rightarrow {A}}$, as we iterate through the set ${\mathcal{X}}$ (${\mathcal{X}}^0_{\backslash P_1}$). To understand the result of this calculation, we first point out that the subsets $\mathcal{P}$ and $\mathcal{P}'$ are defined such that their respective elements combine to provide identical information about the actions (see Equation \ref{eqn:elems_pcr_cond}). Therefore, for a single element, the following holds: 
\begin{equation}
\label{eqn:algo_proof1}
\begin{split}
H({A}|{\mathcal{X}}^0_{\backslash ( \mathcal{P}, \mathcal{P}')} \cup \mathcal{P}') & < H({A}|{\mathcal{X}}^0_{\backslash ( \mathcal{P}, \mathcal{P}')} \cup \mathcal{P}'_{\backslash P'_i}).
\end{split}
\end{equation}
Consequently, the following must also be true:
\begin{equation}
\label{eqn:algo_proof2}
\begin{split}
H({A}|{\mathcal{X}}^0_{\backslash ( \mathcal{P}, \mathcal{P}')} \cup \mathcal{P}' \cup \mathcal{P}_{\backslash P_1}) & < H({A}|{\mathcal{X}}^0_{\backslash ( \mathcal{P}, \mathcal{P}')} \cup \mathcal{P}'_{\backslash P'_1} \cup \mathcal{P}_{\backslash P_1}) \\ 
H({A}|{\mathcal{X}}^0_{\backslash P_1} ) & < H({A}|{\mathcal{X}}^0_{\backslash \{P_1, P'_1\}} ) 
\\ 
H({A}|{\mathcal{X}} ) & < H({A}|{\mathcal{X}}_{\backslash P'_1} ).
\end{split}
\end{equation}
Thus, $\Phi_{P'_1; {\mathcal{X}} \rightarrow {A}} > 0$. We now include $P'_1$ in ${\mathcal{X}}_{A_1}$ and we do not remove it from ${\mathcal{X}}$. 
We will then consider variable $P_2$, and exclude it from ${\mathcal{X}}$ for similar reasons to $P_1$. The variable $P'_2$ will then remain in the set ${\mathcal{X}}$ and be added to ${\mathcal{X}}_{A_1}$ for reasons similar to $P'_1$. This process will be repeated until we have removed all the elements of the subset $\mathcal{P}$ from the set ${\mathcal{X}}$, while the elements of the set $\mathcal{P}'$ have been added to ${\mathcal{X}}_{A_1}$. We have therefore proved that Algorithm \ref{alg:algorithm} reliably addresses cases of CPMCR between subsets with equal cardinality.  $\square$

\noindent \textbf{Theorem 2.} \textit{See Section \ref{sec:prac_app}.}

\noindent \textit{Proof. }Given Condition 1 and the proof of Lemma 5 outlined in Appendix \ref{appendix:alg_correct}, we can disregard cases of CPMCR, and, therefore, directly apply Theorem 1 to complete this proof. $\square$

\section{Tit-For-N-Tats Extra Graphs} \label{appendix:tft_graphs}

\begin{figure*}[ht]
    \centering
    \subfigure(A){\includegraphics[width=14.5cm]{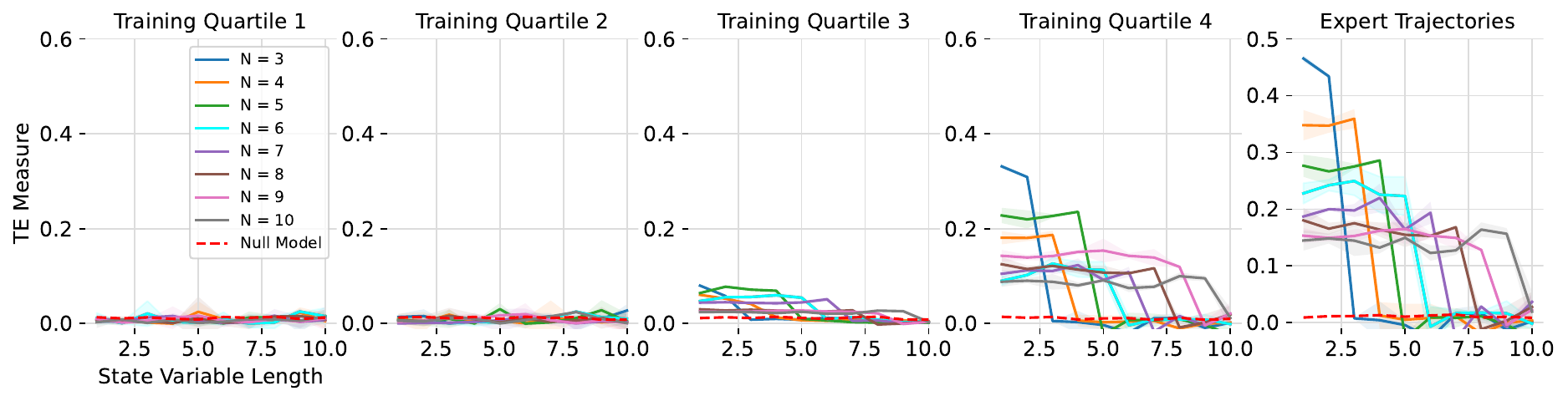} }

    \caption{In subfigure (A), we use $\Phi_{{X}_i;{\mathcal{X}} \rightarrow {A}}$ to investigate how entropy is transferred from observable state variables to actions during different stages of training.   }
    \label{fig:pd_ex}
\end{figure*}

\section{Implementation Details} \label{appendix:impdet}

\subsection{Synthetic Experiments}
\label{appendix:synthhyp}

\paragraph{Transfer entropy measure estimation.} For the estimation of the amount of entropy that each variable transfers to the actions, we use a standard feed-forward Neural Network with a ReLu activation function and with one hidden layer with 50 nodes. We select the following values for the parameters in Algorithm \ref{alg:te_est}: $b=10000$, $N=20000$, with a learning rate of $0.01$. 

\paragraph{Baselines.} For the calculation of UMFI, we select the exact values of the hyperparameters as those used in \citep{janssen2023}.
As far as PI is concerned, we adopt the following values: $\alpha = 0.01$, and 1000 runs. For the calculation of PI itself, we employed Ridge \citep{pedregosa2011}, as implemented in Scickit-Learn. In fact, in this case, hyperparameter tuning has no impact on the final UMFI and PI values.
\subsection{The Secret Key Game}
\paragraph{Trajectory generation.} To carry out the polynomial interpolation in the Secret Key Game, we use the NumPy's \verb+polynomials+ package \url{https://numpy.org/doc/stable/reference/routines.polynomials.package.html}. An expert trajectory for this game is one in which the agent achieves a score of zero. For handling the state spaces of lengths 25 and 50, we employ one-step temporal difference Actor-Critic learning during the execution of the game.

The Actor network is updated using Equation $\theta_{actor}' \gets \theta_{actor} + \alpha_{actor} \gamma^t A \nabla \ln \pi(a^t|s^t, \theta_{actor})$, where $\delta = r + \gamma v(s^t, \theta_{critic}) - v(s^{t+1}, \theta_{critic})$ \citep{sutton2018}. The Critic network is updated using the following equation: $\theta_{critic}' \gets \theta_{critic} + \alpha_{critic} MSE(\delta)$. Both networks are characterized by a single fully connected layer of size 64 with a ReLu  activation function. The other values of the hyperparameters of the networks are $\gamma = 0.99, \alpha_{actor} = 0.0001, \alpha_{critic} = 0.001$.

\paragraph{Transfer entropy measure estimation.} For the estimation of the amount of entropy that each of the keys transfers to the actions, we again use a network with one hidden layer with 50 nodes and a ReLu  activation function. For 25 keys, we use $b=5000$, $N=2000$ in Algorithm \ref{alg:te_est}. Instead, for a state with 50 keys, we require the following parameters for Algorithm \ref{alg:te_est} $b=10000$, $N=20000$. For both, we used a learning rate of $0.01$.

\paragraph{Baselines.} For both 25 and 50 keys, we use the exact hyperparameter values specified in \citep{janssen2023} for calculating UMFI, as they are sufficient to identify the correct keys in the game. For the PI calculation, we employ the same hyperparameter values as those used in the synthetic data experiments.

\paragraph{Evaluation of designed states.} We employ a one-step temporal difference Actor-Critic architecture when evaluating each state. Both networks have a single fully connected layer of size 64 and uses a ReLu  activation function. We employ the following values for the parameters: $\gamma = 0.99, \alpha_{actor} = 0.0001, \alpha_{critic} = 0.001$, considering 20,000 episodes of the game (in this game each episode requires only one iteration).

\subsection{OpenAI's Gym: Cart Pole}
\paragraph{Trajectory generation.} We play the game until convergence. An expert trajectory is defined as one that achieves a cumulative reward greater than 475, in line with the criteria proposed by the authors of the environments \citep{brockman2016openai}. This is done using the same Actor-Critic architecture as described for the Secret Key Game. We introduce random variables into the state, each following a uniform probability distribution within the range $V_{rand_i} \in [-5, 5]$. However, if the variables are truly random, the specific range should be inconsequential.

\paragraph{Transfer entropy measure estimation.} We use the same architecture as previously stated for the Secret Key Game, except in this case, we use the following parameters for Algorithm \ref{alg:te_est} $b=10000$, $N=4000$, and $\alpha = 0.01$.

\paragraph{Baselines.} For these experiments when calculating UMFI  we use the same values of hyperparameters as in \citep{janssen2023}, except for the fact that we adopt 50 trees instead of 100. This was because we saw improved performance using 50 trees for the task at hand. For the PI calculation, we adopted the same values of the hyperparameters as those used for the synthetic data experiments.

\paragraph{Evaluation of designed states.} When plotting the state performance curves shown in Figure \ref{fig:CartPole}, we use the same architecture as for the generation of the trajectories.

\subsection{OpenAI's Gym: Lunar Lander}
\paragraph{Trajectory generation.}
In accordance with Open AI's documentation, we consider expert trajectories as those which achieve a cumulative reward above $200$. Again, the method is applied after playing the game until convergence. This is done using the same Actor-Critic architecture as described for the Secret Key Game. We inject three random variables into the state, in the same manner as described for the cart-pole game. 

\paragraph{Transfer entropy measure estimation.}
We use the same architecture as previously stated, but here we use the following parameters for Algorithm \ref{alg:te_est}: $b=25000$, $N=100000$, and a learning rate of $0.01$.

\paragraph{Baselines.} Again, for these experiments when calculating UMFI,  we adopt the same values of the hyperparameters as used in \citep{janssen2023}, except for the fact that we use 50 trees instead of 100 (since 50 trees are sufficient to separate the informative variables from the non-informative ones). For the PI calculation, we again adopted the same values of the hyperparameters as used for the synthetic data experiments.

\paragraph{Evaluation of designed states.} For the evaluation of the training times corresponding to states composed of different numbers of variables, we use the same architecture as that used for generating trajectories. In this case, we only run the experiment for 3000 episodes. 

\subsection{OpenAI's Gym: Pendulum}
\paragraph{Trajectory generation.}
Again we play the game until convergence. This is done using a PPO architecture, with the following Actor update rule:
\begin{equation}
\begin{split}
     \theta' \gets \theta +   \alpha \gamma^t   \nabla  min(\frac{\pi(a^t|s^t, \theta^t)}{\pi(a^t|s^t, \theta_k^t)}  {A}^t, clip(\frac{\pi(a^t|s^t, \theta^t)}{\pi(a^t|s^t, \theta_k^t)}, 1 - \epsilon, 1 + \epsilon) {A}^t,
\end{split}
\end{equation}
where the surrogate actor had its weights updated as described except it is multiplied by $0.95$. For this algorithm, we use the following values for the hyperparameters: $\gamma = 0.99, \alpha = 0.0003, \epsilon = 0.2$. We also assume a mini-batch size of 64 and update our networks every 2048 steps. We use 10 epochs and an entropy coefficient of 0.001. The network has one hidden layer with 64 nodes and uses continuous hyperbolic tangent activation functions according to \cite{schulman2017}. We consider trajectories with cumulative rewards greater than -10 as expert trajectories. 

\paragraph{Transfer entropy measure estimation.} We use the same architecture to estimate the TE values, combined with the following parameters for Algorithm \ref{alg:te_est} $b=5000$, $N=4000$, with a learning rate of $0.01$.

\paragraph{Baselines.} For these experiments, when calculating UMFI we employ the same values of the hyperparameters as those used in \citep{janssen2023}. For the PI calculation, we adopted the same hyperparameters as used for the synthetic data experiments.

\paragraph{Evaluation of designed states.} We use the same PPO architecture when investigating the performance of different states as that used for generating the trajectories. 

\subsection{TFNT in the Iterated Prisoner's Dilemma}
\label{appendix:pdmat}

\begin{figure}[ht]
\centering
\begin{tikzpicture}[element/.style={minimum width=2.cm,minimum height=.1cm}]
\matrix (m) [matrix of nodes,nodes={element},column sep=-\pgflinewidth, row sep=-\pgflinewidth,]{
         & Cooperate  & Defect  \\
Cooperate & |[draw]|2, 2 & |[draw]|3, 0 \\
Defect & |[draw]|0, 3 & |[draw]|1, 1 \\    };
\node[above=0.25cm] at ($(m-1-2)!0.5!(m-1-3)$){\textbf{Player 1}};
\node[rotate=90] at ($(m-2-1)!0.5!(m-3-1)+(-1.25,0)$){\textbf{Player 2}};
\end{tikzpicture}
\caption{Iterated Prisoner's Dilemma payoff matrix.}
\label{fig:ipd_mat}
\end{figure}

\paragraph{Trajectory generation.}
We employ tabular Q-learning where $\alpha = 0.9$, and the value of $\epsilon$ decays from $1$ to $0$ during $40,000$ iterations ($\epsilon^{t} = \epsilon^{t-1} - (1/40,000) $), while $\gamma$ is equal to $0.99$. We play the game with the above payoff matrix until convergence. We use one-shot encoding in our implementation to generate the state at a certain time step. For example, let us assume that the last 9 action pairs could be represented by $s^t = [(C, D), (C, D),(C, C),(D, D),(C, D),(D, D),(C, D),(D,C),(D,C)] $,  this can be re-written as: $s^t = [(2), (2),(0),(3),(2),(3),(2),(1),(1)] $.

\paragraph{Transfer entropy measure estimation.} We use the same architecture as for the other experiments combined with the following values of the parameters for Algorithm \ref{alg:te_est}: $b=1000$, $N=10000$, with a learning rate of $0.01$. 

\paragraph{Baselines.} For these experiments, for the calculation of UMFI,  we use the exact values of the hyperparameters as those in \citep{janssen2023}. We select these values since hyperparameter tuning has no impact on the final UMFI values. For the PI calculation, we again adopt the same hyperparameters as those used for the synthetic data experiments.

\paragraph{Evaluation of designed states.} We employ the same Q-learning architecture when evaluating the training performance of different state representations.

\subsection{Null Model}
The null model consists of a series of random integers between 0 and 1 that are added to the trajectories after their generation. For all experiments, the values of the hyperparameters used in Algorithm \ref{alg:te_est} to estimate $\Phi_{NM;{\mathcal{X}} \rightarrow {A}}$ are the same as for the variables of interest. For a variable to be considered as influencing the actions in a statistically significant way they must have a $95\%$ chance of falling outside the range of the null model, and, therefore, $\mu_{Xi} - \frac{2\sigma_{Xi}}{\sqrt{10}} > \mu_{NM} + \frac{2\sigma_{NM}}{\sqrt{10}}$, for all variables included in ${\mathcal{X}}_*$.

\end{document}